%% file: main.tex
\PassOptionsToPackage{numbers,sort&compress}{natbib}
\documentclass[11pt,a4paper,logo]{googledeepmind}

\setrightlogo[160pt]{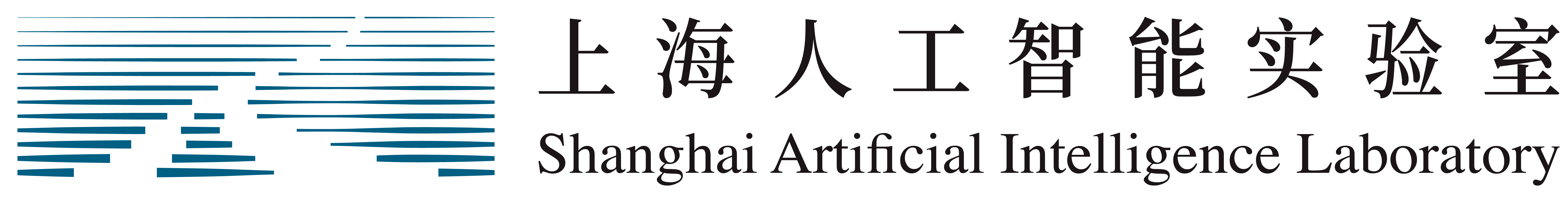}

\usepackage{natbib}
\usepackage{array}
\usepackage{multirow}
\usepackage{listings}
\usepackage{float}
\usepackage[most]{tcolorbox}

\microtypesetup{tracking=false}

\hypersetup{
  colorlinks=true,
  linkcolor=blue,
  citecolor=blue,
  urlcolor=blue,
  pdftitle={SciDataSailor: Deep Scientific Data Exploring},
  pdfauthor={Jiyong Rao, Yicheng Qiu, Chunfeng Song, and Runkai Zhao}
}

\newcolumntype{Y}{>{\centering\arraybackslash}X}

\providecommand{\Description}[1]{}

\newcommand{\github}{%
  \raisebox{-1.5pt}{\includegraphics[height=1em]{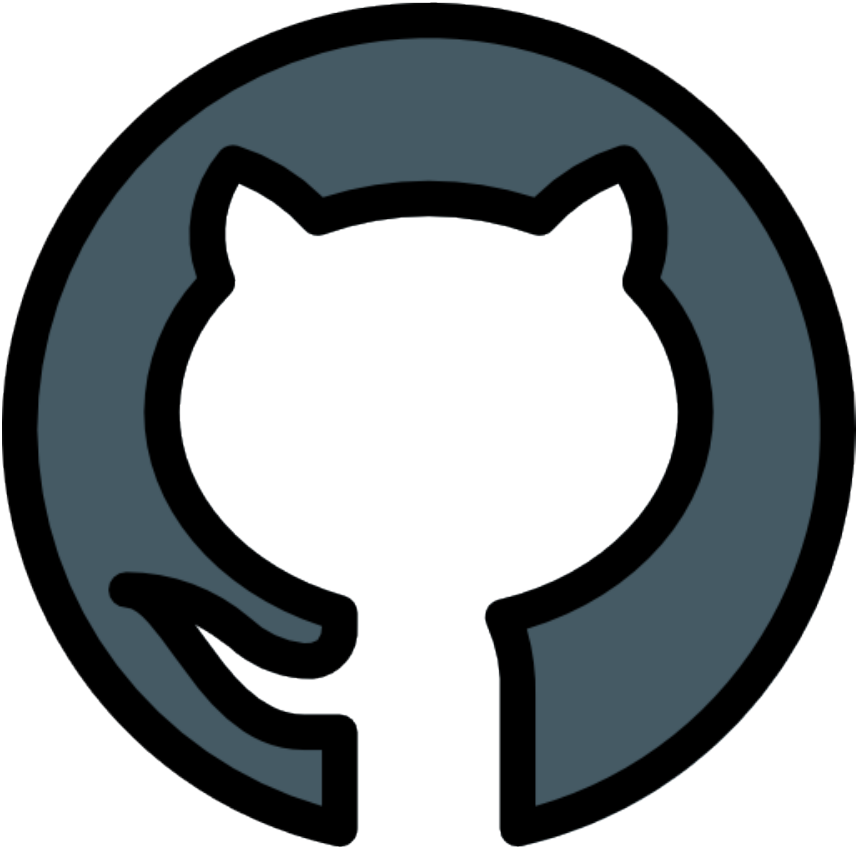}}%
}
\newcommand{\huggingface}{%
  \raisebox{-1.5pt}{\includegraphics[height=1em]{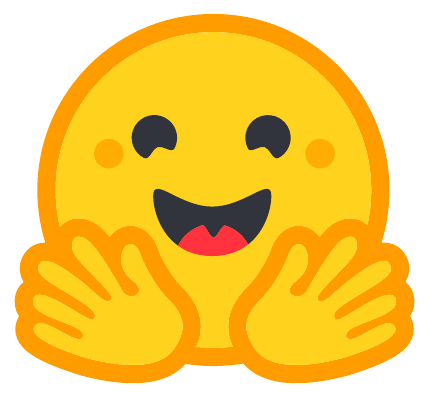}}%
}

\title{%
  \texorpdfstring{%
    \raisebox{-0.1\height}{%
      \includegraphics[height=1em]{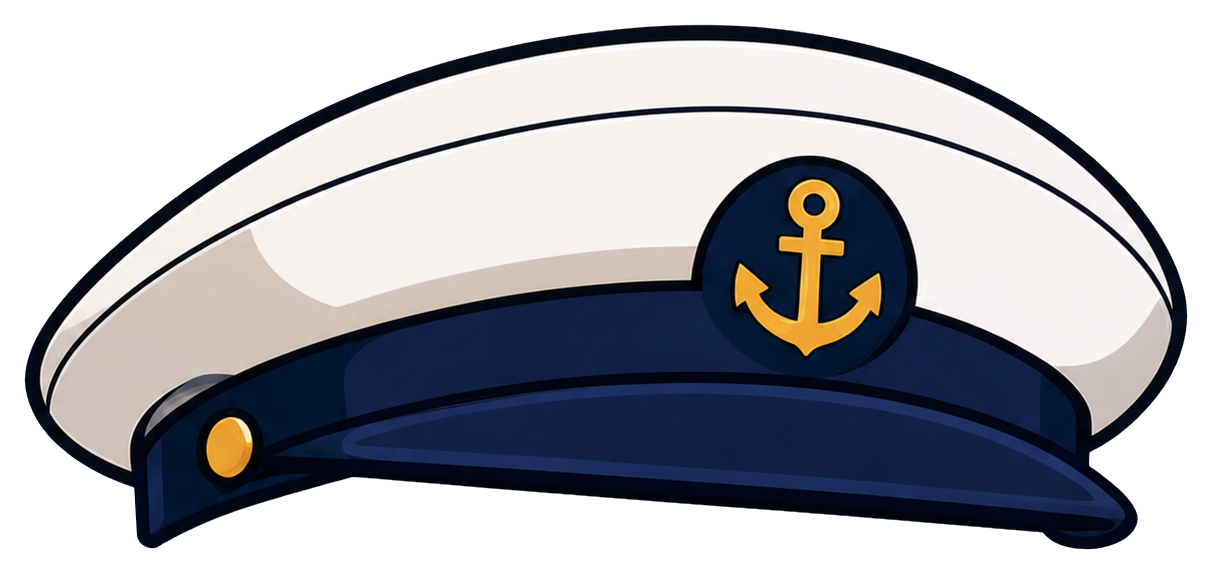}%
    }%
    \hspace{0.35em}SciDataSailor: Deep Scientific Data Exploring%
  }{SciDataSailor: Deep Scientific Data Exploring}%
}
\author{Jiyong Rao*, Yicheng Qiu*, Chunfeng Song\textdagger, and Runkai Zhao\textdagger\ensuremath{\clubsuit}}

\begin{abstract}
\input{components/0-Abstract}
\end{abstract}

\makeatletter
\newcommand{\makeboxedfrontmatter}{%
  \vspace*{-30pt}%
  \begin{tcolorbox}[
    enhanced,
    colback=cyan!5!white,
    colframe=cyan!5!white,
    boxrule=0pt,
    arc=10pt,
    left=18pt,
    right=18pt,
    top=8pt,
    bottom=8pt,
    width=\textwidth
  ]
  {\normalfont\bfseries\fontsize{17}{20}\selectfont
    \raggedright\hyphenpenalty=10000\exhyphenpenalty=10000
    \@title\par}
  \vspace{0.8em}
  {\normalfont\bfseries\fontsize{11}{14}\selectfont
    Jiyong Rao\textsuperscript{1,2*}\quad
    Yicheng Qiu\textsuperscript{1*}\quad
    Chi Zhang\textsuperscript{1}\quad
    Chunfeng Song\textsuperscript{1\textdagger}\quad
    Runkai Zhao\textsuperscript{1\textdagger\ensuremath{\clubsuit}}\par}
  \vspace{0.4em}
  {\normalfont\fontsize{10.5}{13}\selectfont
    \textsuperscript{1}Shanghai Artificial Intelligence Laboratory\quad
    \textsuperscript{2}Tongji University\par}
  {\large*} Equal contribution.\quad
  {\large \textdagger} Corresponding author.\quad
  {\large $\clubsuit$} Project lead.\par
  \vspace{1.1em}
  {\normalfont\fontsize{11}{13.5}\selectfont
    \theabstract\par}
  \vspace{0.9em}
  {\normalfont\fontsize{10.5}{13}\selectfont
    \github\ \textbf{Data Curation Code:}
    \href{https://github.com/SciDataOcean/SciDataSailor}
      {\texttt{SciDataOcean/SciDataSailor}}\\
    \github\ \textbf{Evaluation Scaffolding:}
    \href{https://github.com/SciDataOcean/simple-agent-eval-scaffolding}
      {\texttt{SciDataOcean/simple-agent-eval-scaffolding}}\par}
    \huggingface\ \textbf{Verified Agent Trajectory:}
    \href{https://huggingface.co/datasets/SciDataOcean/SciDataSailor-SFT-2K}
      {\texttt{SciDataOcean/SciDataSailor-SFT-2K}}\\
  \end{tcolorbox}
}
\makeatother

\titleformat{\section}
  {\normalfont\bfseries\fontsize{20}{24}\selectfont}
  {\thesection}
  {1em}
  {#1}
  []
\titlespacing*{\section}
  {0pt}
  {2.2ex plus 4pt minus 2pt}
  {1.2ex plus 2pt minus 1pt}

\begin{document}

\makeboxedfrontmatter
\thispagestyle{firststyle}

\begin{figure}[H]
  \vspace{5mm}
  \centering
  \includegraphics[width=\textwidth]{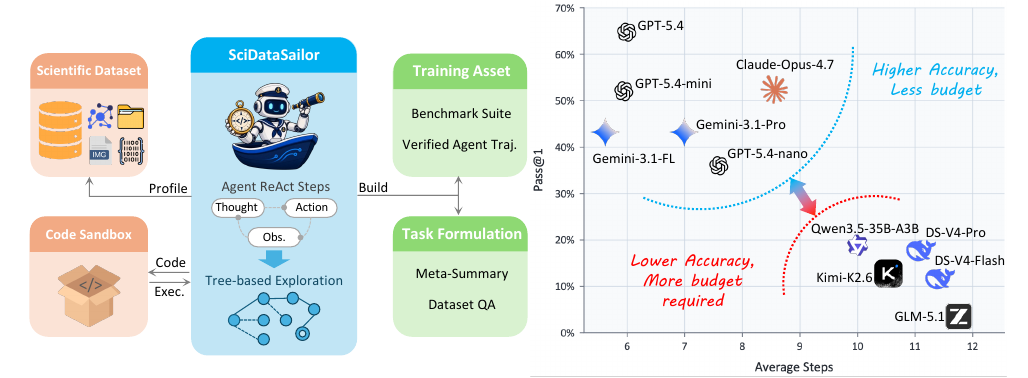}
  \caption{\textbf{Overview and behavioral gap in deep scientific data
  exploring.} Left: \textit{SciDataSailor} constructs training and benchmark
  assets through scientific-database and code-sandbox interaction. Right:
  under a 12-step budget, proprietary models achieve higher Pass@1 with
  fewer ReAct steps than open-weight models. Fine-tuning Qwen3.5-9B on
  \textit{SciDataSailor-SFT-2K} improves Pass@1 from 14.01 to 28.99 and
  success rate from 49.76 to 96.14 while reducing average steps from 10.21
  to 6.24.}
  \Description{A pipeline diagram shows the construction of SciDataSailor
  training and benchmark assets. A scatter plot below compares proprietary
  and open-weight agents by Pass@1 and average ReAct steps, with proprietary
  agents concentrated toward higher accuracy and fewer steps.}
  \label{fig:behavior_gap}
\end{figure}

\input{components/1-Introduction}

\input{components/2-LiteratureReview}
\input{components/3-Method}
\input{components/4-Experiments}
\input{components/5-Conclusion}

\clearpage
\begingroup
\small
\sloppy
\bibliographystyle{unsrtnat}
\bibliography{main}
\endgroup

\clearpage
\appendix
\input{components/X-appendix.tex}

\end{document}

%% file: components/0-Abstract.tex
Scientific datasets are commonly organized as hierarchical repositories containing heterogeneous and interdependent files, making their inspection, integration, and analysis labor-intensive and reliant on domain expertise.
Although large language model (LLM) agents have advanced substantially in planning, reasoning, and tool use, existing research has largely overlooked their ability to interact with real scientific data assets through executable environments.
We introduce \emph{deep scientific data exploration}, an agentic task paradigm in which agents navigate repositories, interpret heterogeneous files and schemas, execute analyses, integrate cross-file evidence, and produce conclusions grounded in executed observations.
To operationalize this paradigm, we present \textit{SciDataSailor}, a framework for synthesizing tool-interactive trajectories by balancing broad exploration with targeted exploitation.
\textit{SciDataSailor} instantiates trajectory synthesis as Monte Carlo Tree Search (MCTS) with four task-specific mechanisms: difficulty-stratified exploration seeds, dual-feedback first-play urgency, hierarchical strategy-to-tool action generation, and entropy-guided branching.
Using this framework, we construct \textit{SciDataSailor-SFT-2K} for supervised fine-tuning and \textit{SciDataSailor-Bench} for evaluation, with the latter comprising 627 meta-information summarization tasks and 586 scientific question-answering tasks across 27 datasets spanning the life, earth, and physical sciences.

%% file: components/1-Introduction.tex
\section{Introduction}

Scientific discovery increasingly depends on data stored in large, hierarchical repositories containing heterogeneous and interdependent files~\cite{fayyad1997data,rao2026scidatacopilot,alber2026cellvoyager,youngblut2025scbasecount}.
Unlike natural-language corpora, which encode knowledge primarily in passages and expose explicit schemas through standardized query interfaces, scientific repositories distribute evidence across raw measurements, metadata, annotations, experimental records, and derived artifacts.
Their files may differ in modality, format, data organization, identifier conventions, units, and coordinate systems, while the interpretation of one artifact often depends on information stored elsewhere in the repository~\cite{kumar2021data,hendler2014data,yu2024does,gong2022repository}.
Recovering reliable evidence therefore requires researchers to navigate directories, interpret implicit schemas, align related artifacts, compute descriptive evidence, and validate conclusions, which is a labor-intensive and error-prone process that scales poorly to large or sparsely documented datasets.

Large language model (LLM) agents have advanced substantially in planning, reasoning, and tool use, but most existing agentic systems operate over web documents and textual corpora by accessing structured query interfaces~\cite{tan2026hiersearch,li2025towards,li2025survey,cen2025sqlfixagent,xia2025r3,zhang2025web,huang2025deep}.
To address the comparatively understudied problem of interacting directly with real scientific data assets in executable environments, we introduce \textbf{\emph{Deep Scientific Data Exploring}}, an agentic task paradigm in which an agent navigates an unfamiliar repository, inspects heterogeneous files and schemas, executes analyses, connects evidence across data assets, and produces conclusions grounded in observed tool outputs.
This setting moves beyond retrieving relevant text: the agent must determine what evidence is available, decide which operations to execute, maintain state across intermediate observations, and adapt its exploration as the repository structure and data semantics are progressively revealed.

Deep scientific data exploring requires balancing exploration and exploitation over long horizons.
Broad exploration is necessary to discover file organization, modalities, variables, and cross-file relationships, whereas targeted exploitation is required to verify promising findings, compute precise statistics, and assemble evidence for a final response.
A successful trajectory may involve file-system navigation, format inspection, schema inference, metadata alignment, statistical profiling, and claim validation, thereby providing process-level supervision that records how evidence is discovered, tested, and integrated.
However, manually constructing diverse and reliable trajectories over heterogeneous repositories is prohibitively expensive.

Existing trajectory-synthesis pipelines are primarily designed for deep website research~\cite{wu2026deepresearch,lan2025deepwidesearch}, where agents iteratively reformulate queries, retrieve documents, read passages, and synthesize answers~\cite{du2026openseeker,li2025websailor-v1,li2025websailor-v2,huang2025deep,zhang2025web,alzubi2025open}.
Such pipelines do not directly capture the executable and provenance-sensitive operations required for scientific data exploration.
Because scientific data trajectories must remain faithful to file contents and computed observations while preserving cross-artifact dependencies, a dedicated framework is needed to balance search breadth and depth and reject unsupported traces.

\begin{figure}[t]
  \centering
  \includegraphics[width=\textwidth]{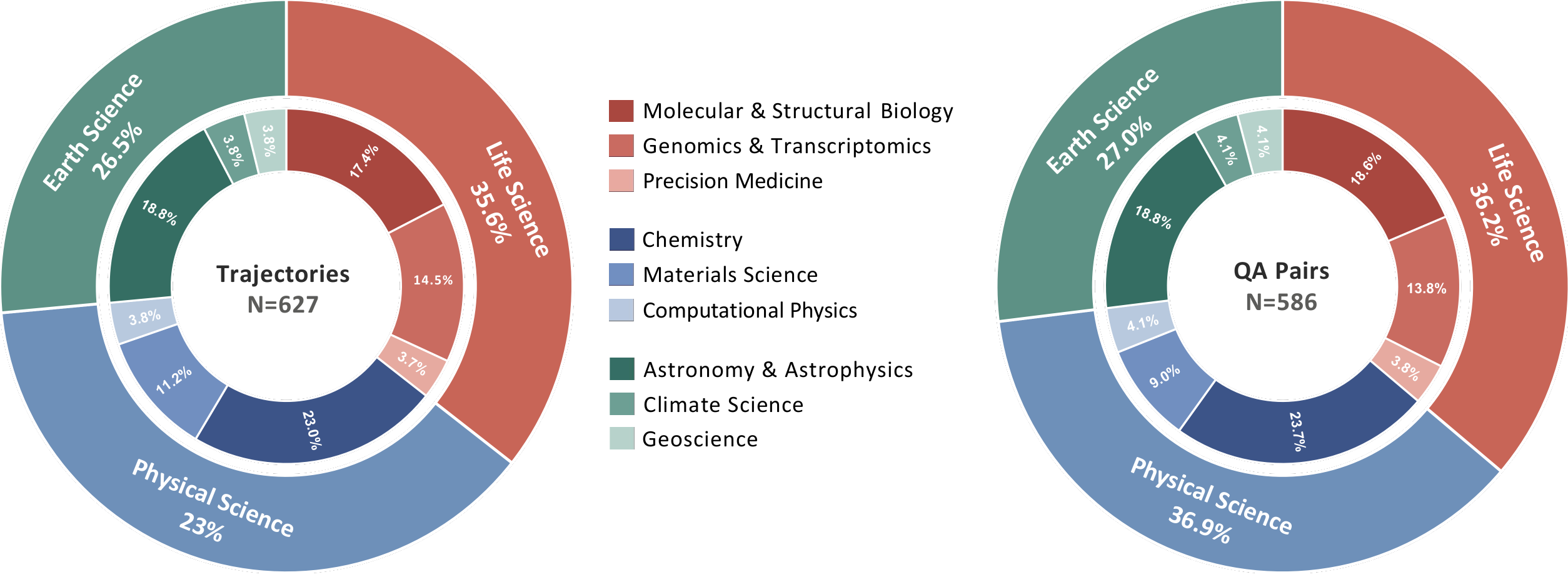}
  \caption{\textbf{Dataset taxonomy and task composition of \textit{SciDataSailor-Bench}.} The benchmark covers 27 heterogeneous datasets from the life, earth, and physical sciences, comprising 627 meta-information summarization tasks and 586 evidence-grounded scientific question-answering tasks.}
  \Description{A taxonomy diagram groups 27 scientific datasets into the life, earth, and physical sciences and shows two benchmark tracks containing 627 meta-information summarization tasks and 586 scientific question-answering tasks.}
  \label{fig:benchmark_overview}
\end{figure}

To address this need, we present \textit{SciDataSailor}, a framework for synthesizing tool-interactive scientific data trajectories in realistic executable environments.
\textit{SciDataSailor} adapts Monte Carlo Tree Search (MCTS) with evidence-seeking designs to balance broad exploration with targeted goals, using execution feedback to focus computation on the most informative trajectories. As follows,
Execution-validity and hallucination checks is incorporated to remove malformed, uninformative, or unsupported trajectories. Using this framework, we construct \textit{SciDataSailor-Bench} and \textit{SciDataSailor-SFT-2K}.
The benchmark evaluates agents on 27 raw datasets spanning the life, earth, and physical sciences, with 627 meta-information summarization tasks and 586 scientific QA tasks, as summarized in Figure~\ref{fig:benchmark_overview}. Compared with existing scientific-domain benchmarks, \textit{SciDataSailor-Bench} provides broader coverage across six dimensions, as highlighted in Table~\ref{tab:benchmark_comparison}.
The supervised fine-tuning (SFT) corpus is designed to train scientific data agents using tool-calling exploration trajectories, and its composition is detailed in Figure~\ref{fig:supp_sft_2k} in Appendix~\ref{app:sft-2k-overview}.
We conduct extensive comparisons among proprietary and open-weight model variants under a unified executable-agent protocol.
The comparison reveals a marked behavioral gap: proprietary models generally explore more consistently and efficiently, whereas open-weight models often require additional trial-and-error turns to reach successful outcomes, as illustrated in Figure~\ref{fig:behavior_gap}.
Moreover, token-level supervision over our verified agentic trajectories transfers effective environment-interaction and evidence-seeking strategies to a compact open-weight model, improving both accuracy and efficiency. Our main contributions are as follows:
\begin{itemize}
    \item \textbf{A task formulation and synthesis methodology for scientific data exploration.}
    We formulate \emph{deep scientific data exploring} as long-horizon, tool-calling interaction with real scientific repositories and introduce \textit{SciDataSailor}, which synthesizes executable and evidence-grounded trajectories through difficulty-stratified seeds, dual-feedback tree search, hierarchical action generation, and uncertainty-adaptive branching.

    \item \textbf{Executable training and evaluation resources.}
    We curate \textit{SciDataSailor-SFT-2K} for trajectory-level supervised fine-tuning and \textit{SciDataSailor-Bench} for evaluation. The benchmark comprises 627 meta-information summarization tasks and 586 scientific QA tasks across 27 datasets spanning the life, earth, and physical sciences.

    \item \textbf{An empirical characterization of scientific data agents.}
    Under a unified executable protocol, we identify systematic differences in the reliability and efficiency of proprietary and open-weight agents and show that token-level supervision substantially improves the accuracy, completion rate, and interaction efficiency of a compact open-weight model, particularly under constrained ReAct-step budgets.
\end{itemize}

%% file: components/2-LiteratureReview.tex
\section{Literature Review}
\label{sec:related_work}

\begin{table}[t]
\centering
\def\cmark{\textcolor{green!60!black}{\ding{51}}}
\def\xmark{\textcolor{red!75!black}{\ding{55}}}
\def\pmark{\textcolor{orange!85!black}{\ding{115}}}
\def\ours#1{\cellcolor{green!12}#1}
\caption{\textbf{Comparison with scientific-domain benchmarks.} A green checkmark \cmark\ denotes explicit support, an orange triangle \pmark\ denotes partial support, and a red cross \xmark\ denotes no support.}
\label{tab:benchmark_comparison}
\footnotesize
\setlength{\tabcolsep}{2pt}
\renewcommand{\arraystretch}{1.15}
\begingroup
\renewcommand{\tabularxcolumn}[1]{m{#1}}
\begin{tabularx}{\textwidth}{@{}>{\raggedright\arraybackslash}m{0.17\textwidth}YYYYYYY@{}}
\toprule
\multirow{2}{*}{\textbf{Dimension}} &
\shortstack{\textbf{LAB}\\\textbf{Bench}} &
\shortstack{\textbf{BioKG}\\\textbf{Bench}} &
\textbf{CURIE} &
\shortstack{\textbf{Science}\\\textbf{AgentBench}} &
\textbf{BixBench} &
\shortstack{\textbf{BioAgent}\\\textbf{Bench}} &
\ours{\shortstack{\textbf{SciData}\\\textbf{Sailor-Bench}}} \\
& \cite{laurent2024labbench}
& \cite{lin2024biokgbench}
& \cite{cui2025curie}
& \cite{chen2024scienceagentbench}
& \cite{mitchener2025bixbench}
& \cite{fa2026bioagentbench}
& \ours{\textbf{(Ours)}} \\
\midrule
Scientific-domain diversity & \xmark & \xmark & \cmark & \cmark & \xmark & \xmark & \ours{\cmark} \\
Data-format heterogeneity & \pmark & \pmark & \xmark & \pmark & \pmark & \cmark & \ours{\cmark} \\
Open-ended exploration & \xmark & \xmark & \xmark & \pmark & \pmark & \pmark & \ours{\cmark} \\
Executable tool support & \pmark & \pmark & \xmark & \cmark & \cmark & \cmark & \ours{\cmark} \\
Raw-repository grounding & \cmark & \cmark & \cmark & \pmark & \pmark & \pmark & \ours{\cmark} \\
Training trajectories & \xmark & \xmark & \xmark & \xmark & \xmark & \xmark & \ours{\cmark} \\
\bottomrule
\end{tabularx}
\endgroup
\end{table}

\subsection{Agentic Scientific Data Benchmarking}
Table~\ref{tab:benchmark_comparison} compares \textit{SciDataSailor-Bench} with representative scientific-domain benchmarks.
Recent benchmarks have begun to evaluate LLMs and agents beyond static, closed-form QA.
LAB-Bench measures biology-research capabilities across literature, protocols, and domain reasoning tasks~\cite{laurent2024labbench}, while BioKGBench evaluates biomedical knowledge-graph checking~\cite{lin2024biokgbench}.
CURIE targets multitask scientific long-context understanding and reasoning across multiple scientific domains~\cite{cui2025curie}.
ScienceAgentBench studies language agents for data-driven scientific discovery~\cite{chen2024scienceagentbench}, and BixBench and BioAgent Bench further evaluate agentic workflows in computational biology and bioinformatics pipelines~\cite{mitchener2025bixbench,fa2026bioagentbench}.
Beyond benchmark construction, MLEvolve investigates automated machine-learning algorithm discovery, while Intern-Atlas organizes methodological evolution as research infrastructure for AI scientists~\cite{du2026mlevolve,wu2026intern}.
These benchmarks broaden the evaluation of LLM agents over documents, knowledge graphs, code, and domain workflows, but rarely capture raw scientific repositories where agents must inspect files, infer schemas, align metadata, execute code, and ground conclusions in observed evidence.
Repository exploration has also emerged as an evaluation target in software engineering: SWE-Explore examines how coding agents navigate unfamiliar code repositories, a neighboring setting that likewise emphasizes information discovery over hierarchical artifacts~\cite{zhang2026swe}.
\textit{SciDataSailor} therefore targets \textbf{scientific data exploration} and evaluates both meta-information summarization and scientific QA tasks.

\subsection{Tool-Using Agents and Trajectory Synthesis}
Recent data-agent systems increasingly use executable tools, code rollouts, and synthetic trajectories~\cite{qiao2026datamind,rao2026scidatacopilot,dong2026agentic,zhang2025deepanalyze}.
DataMind~\cite{qiao2026datamind} studies scalable data synthesis and training recipes for generalist data-analytic agents, while SciDataCopilot~\cite{rao2026scidatacopilot} focuses on agentic data preparation for scientific discovery.
Related systems broaden this direction through unified data preparation and workflow automation in DataFlow~\cite{liang2025dataflow} and data-centric dynamic training in DataFlex~\cite{liang2026dataflex}.
However, scientific data exploration requires process-level supervision for open-ended evidence discovery, not only task-level success on pre-specified objectives.
AgentFlow provides a unified framework for synthesizing agent data~\cite{omniagentsynth2026}. Related work shows that scaling agent-training horizons and data, rather than model parameters alone, can enable a 35B agent to approach the performance of substantially larger models~\cite{bai2026scaling}.
Tree-search agents further improve long-horizon planning and tool use through interactive MCTS~\cite{liang2026imcts}, tree-search reinforcement learning~\cite{hou2026treerl}, and dual-feedback tool planning~\cite{yang2026tooltree, masterman2024landscape, zhu2025menti}, showing the value of exploring and evaluating multiple candidate action paths.
For scientific data exploration, the central bottleneck is not only solving individual tasks but obtaining long-horizon supervision that records how agents discover available evidence, choose among plausible probes, verify intermediate findings, and connect observations to downstream questions.
This motivates a dedicated trajectory synthesis pipeline that can generate diverse, executable, and evidence-rich exploration traces at scale.

%% file: components/3-Method.tex
\section{Preliminaries}
\label{sec:preliminaries}

\begin{figure}[t]
  \centering
  \includegraphics[width=\textwidth]{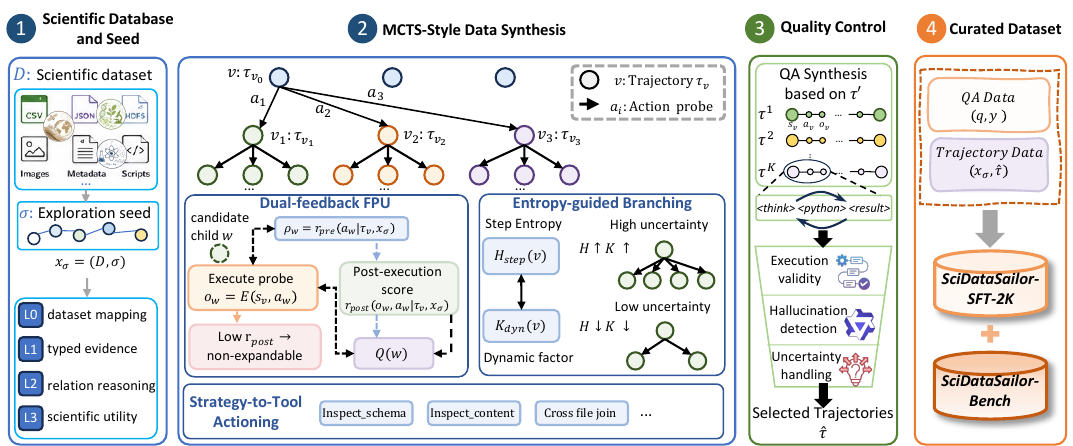}
  \caption{\textbf{Overview of the \textit{SciDataSailor} data synthesis framework.} From scientific datasets and exploration seeds, \textit{SciDataSailor} builds difficulty-stratified, tool-interactive trajectory trees using dual-feedback first-play urgency, entropy-adaptive branching, and hierarchical strategy-to-tool action generation. Filtered trajectories are converted into trajectory and QA data for \textit{SciDataSailor-SFT-2K} and \textit{SciDataSailor-Bench}.}
  \Description{A four-stage framework begins with a scientific dataset and an L0-to-L3 exploration seed, grows an MCTS trajectory tree using dual-feedback first-play urgency and entropy-guided branching, filters trajectories through execution and hallucination checks, and produces the SciDataSailor training and benchmark datasets.}
  \label{fig:framework}
\end{figure}

\noindent\textbf{Long-Horizon Exploration Trajectory.}
We consider scientific data exploration as an interactive process~\cite{yao2023react} over the unknown scientific database $D$.
We define an exploration request as $x=(D,\sigma)$, where $\sigma$ is an exploration seed, and let $\mathcal{E}$ denote the executable code sandbox.
For $t=1,\ldots,T$, the interaction satisfies $a_t\in\mathcal{A}(s_{t-1})$, $o_t=\mathcal{E}(s_{t-1},a_t)$, and $s_t=F(s_{t-1},a_t,o_t)$, where $F$ is the state-transition function.
A completed exploration process is represented as an executable trajectory
\begin{equation}
\tau=(s_0,a_1,o_1,s_1,\ldots,a_T,o_T,s_T,y),
\end{equation}
where $s_0$ encodes the initial request and database context, and $y$ is a final response grounded in the executed observations.
A high-quality trajectory should therefore expose reusable scientific facts about $D$, including schemas, variable distributions, file relationships, numerical summaries, and evidence supporting the final answer.

\noindent\textbf{Tree-Structured Trajectory Space.}
Scientific data exploration naturally admits a tree-structured search formulation because an evidence state often permits multiple feasible next probes.
In this formulation, the root corresponds to the initial state $s_0$, and each node represents a partially executed exploration trajectory.
Sibling nodes encode alternative next actions conditioned on the same evidence state.
Monte Carlo Tree Search (MCTS)~\cite{van2007mcts} grows the search tree by repeatedly applying four stages: \emph{selection}, \emph{expansion}, \emph{simulation}, and \emph{backpropagation}.
Each node $v$ maintains a visit count $N(v)$ and a value estimate $Q(v)$.
During \emph{selection}, MCTS starts at the root and repeatedly chooses the child $v$ of the current node $u$ with the highest exploration--exploitation score, continuing until it reaches an expandable node $u_L$:
\begin{equation}
\label{equ:uct}
\operatorname{UCT}(u,v)=Q(v)+\lambda\sqrt{\frac{\log(1+N(u))}{1+N(v)}}.
\end{equation}
The first term favors children with high estimated utility, whereas the second encourages exploration of less-visited children; $\lambda$ controls the balance between these objectives.
During \emph{expansion}, MCTS executes an untried action $a\in\mathcal{A}(s_{u_L})$, obtains the observation $o=\mathcal{E}(s_{u_L},a)$, and adds the resulting state as a new child $v$.
The \emph{simulation} stage then estimates the utility $\widehat{U}_v$ of the trajectory ending at $v$, either through a rollout or by directly evaluating the executed observations.
Finally, \emph{backpropagation} updates each $z\in\operatorname{Path}(\mathrm{root},v)$, with unmarked and $+$ quantities denoting pre- and post-update values, respectively:
\begin{equation}
N^{+}(z)=N(z)+1,\qquad
Q^{+}(z)=Q(z)+\frac{\widehat{U}_v-Q(z)}{N^{+}(z)}.
\end{equation}
However, vanilla MCTS remains mismatched with our task: its value estimates favor task success over observational richness and diversity, redundant probes can consume the available budget, and retaining only the best leaf is insufficient for data generation.

\section{Probing Scientific Data Exploration Trajectories}
\label{sec:method}
\subsection{Trajectory Synthesis Pipeline}
\label{sec:synthesis_pipeline}
This section describes how \textit{SciDataSailor} turns broad scientific exploration seeds into executable, evidence-grounded trajectories.
Figure~\ref{fig:framework} summarizes the four coupled components of the pipeline.
First, we define scientific data exploration seeds as exploration intents, assign the eight seed intents to executable difficulty levels, and enforce that the synthesized final answer remains anchored to the original intent.
Then, we grow a trajectory tree with Dual-feedback First-Play Urgency (DF-FPU), which uses pre-execution judgments to prioritize promising probes and post-execution observations to calibrate those judgments.
Next, we expand each selected state through hierarchical strategy-to-tool action generation, so that the tree diversifies scientific intents before materializing them as tool probes.
To search effectively, we allocate the branching budget with entropy-adaptive branching, widening uncertain states and narrowing obvious ones.
Together, these components form a closed loop from seed specification to trajectory generation.

\noindent\textbf{Scientific Trajectory Seeds.} \textit{SciDataSailor} starts from scientific trajectory seeds rather than fully specified questions.
A seed $\sigma$ is a broad exploration intent over a scientific dataset, and we assign it to one of four difficulty levels $\ell(\sigma)\in\{\mathrm{L0},\mathrm{L1},\mathrm{L2},\mathrm{L3}\}$ according to the evidence and reasoning needed to answer questions derived from that intent.
The detailed level definitions and the current eight-seed assignment are provided in Appendix~\ref{app:seed_levels}.

\noindent\textbf{Dual-feedback First-Play Urgency.}
Given a scientific data exploration request \(x_\sigma\), \textit{SciDataSailor} incrementally constructs an MCTS trajectory tree. Each node \(v\) represents an executed trajectory prefix \(\tau_v\) and its resulting exploration state \(s_v\). A candidate child \(w\in\operatorname{Ch}(v)\) produces an executable probe \(a_w\in\mathcal{A}(s_v)\). Because \(Q(w)\) is unavailable before its first visit, conventional MCTS treats unvisited children uniformly, which can waste tool budget and disrupt coherent evidence gathering. To address this limitation, we follow ToolTree~\cite{yang2026tooltree} and introduce \emph{Dual-feedback First-Play Urgency} (DF-FPU) to combine a pre-execution estimate of a probe's potential utility with post-execution evidence about its realized utility. Before executing \(a_w\), a pre-execution LLM-Judge assigns
\begin{equation}
\rho_w
=
r_{\mathrm{pre}}
\left(
a_w
\mid
\tau_v,x_\sigma
\right)
\in[0,1],
\end{equation}
where a larger \(\rho_w\) indicates that the probe is more likely to produce relevant and reusable evidence, prioritizing promising first visits. After \(a_w\) is executed, the environment returns \(o_w\) and a post-execution LLM-Judge evaluates its realized evidential utility:
\begin{equation}
y_w
=
r_{\mathrm{post}}
\left(
o_w,a_w
\mid
\tau_v,x_\sigma
\right)
\in[0,1].
\end{equation}
The post-execution score rewards concrete, verifiable evidence and penalizes irrelevant or erroneous outputs. Its return is back-propagated through the selected trajectory, yielding the empirical node value
\begin{equation}
Q(w)
=
\frac{1}{N(w)}
\sum_{i=1}^{N(w)}G_i(w).
\end{equation}
Here, \(G_i(w)\) is either the immediate score \(y_w\) or an aggregate downstream return. Nodes with \(y_w<\theta_{\mathrm{post}}\) remain in the tree for provenance but are marked non-expandable, preventing low-evidence branches from consuming further tool budget.

Although \(\rho_w\) provides a useful prior for an unvisited child, an LLM-Judge may be systematically optimistic or pessimistic in a particular exploration context. DF-FPU therefore calibrates pre-execution scores using the outcomes of previously executed sibling probes. Let
\begin{equation}
\mathcal{S}_v
=
\left\{
u\in\operatorname{Ch}(v)
:
N(u)>0
\right\}
\end{equation}
denote the visited children of \(v\), including children subsequently marked non-expandable. We define the local calibration residual as
\begin{equation}
\Delta_v
=
\begin{cases}
\displaystyle
\frac{1}{|\mathcal{S}_v|}
\sum_{u\in\mathcal{S}_v}
\left(
Q(u)-\rho_u
\right),
&
|\mathcal{S}_v|>0,
\\[1em]
0,
&
|\mathcal{S}_v|=0.
\end{cases}
\end{equation}
The residual \(Q(u)-\rho_u\) captures local prediction error, so \(\Delta_v<0\) corrects overoptimism and \(\Delta_v>0\) corrects underestimation as judge bias varies across exploration states. Using \(\Delta_v\), we define the value used during selection as
\begin{equation}
\widehat{Q}(w)
=
\begin{cases}
Q(w),
&
N(w)>0,
\\[0.4em]
\displaystyle
\operatorname{clip}_{[0,1]}
\left(
\rho_w+\Delta_v-\delta_{\mathrm{fpu}}
\right),
&
N(w)=0,
\end{cases}
\end{equation}
where
\begin{equation}
\operatorname{clip}_{[0,1]}(z)
=
\min\!\left\{1,\max\{z,0\}\right\},
\end{equation}
and \(\delta_{\mathrm{fpu}}\geq0\) is a conservative first-play reduction. For a visited child, DF-FPU uses its empirical value \(Q(w)\). For an unvisited child, it begins with the predicted utility \(\rho_w\), corrects this prediction using the average error observed among its visited siblings, and uses \(\delta_{\mathrm{fpu}}\) to account for the uncertainty associated with an untested action. 



During selection, DF-FPU scores each expandable edge \((v,w)\in\mathcal{F}_{\mathrm{exp}}\) as
\begin{equation}
\operatorname{UCT}_{\mathrm{DF\text{-}FPU}}(v,w)
=
\widehat{Q}(w)
+
\lambda\rho_w
\sqrt{
\frac{\log(1+N(v))}
{1+N(w)}
}
+
\beta d_w,
\end{equation}
where \(\lambda\) controls quality-aware exploration, while the depth bonus \(\beta d_w\), with normalized depth \(d_w\in[0,1]\) and weight \(\beta\geq0\), favors promising multi-step trajectories when frontier nodes lie at different depths. DF-FPU enables evidence-aware exploration by using pre-execution estimates to prioritize first visits, observed outcomes to calibrate node values, and conservative initialization and filtering to limit uncertain or uninformative branches, thereby promoting coherent evidence chains while retaining plausible alternatives.


\noindent\textbf{Hierarchical Strategy-to-Tool Action Generation.}
After DF-FPU selects an expandable node, \emph{Hierarchical Strategy-to-Tool Action Generation} determines what evidence to collect before instantiating the decision as an executable probe.
Its strategy catalog $\mathcal{A}_{\mathrm{str}}$ spans repository and schema inspection, statistical analysis, cross-file integration, claim validation, visualization, and debugging.
At node $v$, the method proposes and scores strategies conditioned on $\tau_v$ and $x_\sigma$, filters weak candidates, converts retained strategies into probes $c_i$, and deduplicates them by tool identity while prioritizing diversity.
This hierarchy reduces redundant probes, promotes complementary evidence gathering, and provides intermediate action-type supervision for SFT/RL credit assignment.

\noindent\textbf{Entropy-Guided Branching.}
We introduce \emph{Entropy-Guided Branching} as an adaptive expansion mechanism, distinct from blocking low-$r_{\mathrm{post}}$ nodes only after execution.
Let $\mathcal{C}_v=\{(c_i,\rho_i)\}_{i=1}^{\bar{M}_v}$ be the probes at node $v$.
For $\bar{M}_v>1$, we compute prior entropy from prospective scores using temperature $T_p>0$:
\begin{equation}
\pi_i=\frac{\exp(\rho_i/T_p)}
{\sum_{j=1}^{\bar{M}_v}\exp(\rho_j/T_p)},\qquad
H_{\mathrm{prior}}(v)=
-\frac{1}{\log \bar{M}_v}\sum_{i=1}^{\bar{M}_v}\pi_i\log\pi_i ,
\end{equation}
with $H_{\mathrm{prior}}(v)=0$ when $\bar{M}_v\leq1$.
When token-level log probabilities are available, $H_{\mathrm{tok}}(v)\in[0,1]$ denotes normalized predictive token entropy. We combine it with prior entropy using nonnegative weights $w_p$ and $w_t$ satisfying $w_p+w_t>0$:
\begin{equation}
\begin{aligned}
H_{\mathrm{step}}(v)&=\frac{w_p H_{\mathrm{prior}}(v)+w_t H_{\mathrm{tok}}(v)}{w_p+w_t},\\
k_{\mathrm{dyn}}(v)&=\min\left(\bar{M}_v,\operatorname{round}\left(k_{\min}+(k_{\max}-k_{\min})H_{\mathrm{step}}(v)\right)\right),
\end{aligned}
\end{equation}
Here, $k_{\min}$ and $k_{\max}$ are the minimum and maximum branching widths. When token entropy is unavailable, we set $H_{\mathrm{step}}(v)=H_{\mathrm{prior}}(v)$.
The diversity filter materializes $k_{\mathrm{dyn}}(v)$ children, while the proposal temperature increases across rollouts to shift from stable early growth to later strategy discovery.
Together with post-pruning after back-propagation, this forms a forward dynamic expansion plus backward quality pruning controller, turning expansion width into a state-dependent variable that improves branch-budget utilization and reduces invalid executions.

\noindent\textbf{Trajectory selection and QA synthesis.}
The steps of trajectories are wrapped by \texttt{<think>}, \texttt{<python>}, \texttt{<result>}, and \texttt{<answer>} tags.
For each trajectory, we identify a compact evidence target from its observations, such as a discovered schema property, a computed statistic, a cross-file linkage, or a data-quality finding.
We then formulate a question $q$ that is entailed by the seed, and set the answer $y$ to the conclusion supported by the trajectory.
The resulting benchmark instance extract the synthesized QA pair $(q,y)$ from the trajectory of meta-information summarization $(x_\sigma,\tau)$. 

\subsection{Quality Control}
\label{sec:quality}
We perfom quality check to ensure that each retained trajectory is executable, evidence-bearing, and free of unsupported scientific claims.
Given a synthesized instance $(x_\sigma,\tau,q,y)$, we verify both the trajectory data $(x_\sigma,\tau)$ and the QA data $(q,y)$.
The central criterion is hallucination detection: every factual statement in the final answer and every important intermediate reasoning step should be recoverable from executed observations rather than from domain priors or internalized knwoledge of LLMs.

\noindent\textbf{Execution validity.} We apply rule-based validation to each trajectory trace.
A trajectory is rejected if it contains unresolved execution errors, missing \texttt{<python>} or \texttt{<result>} blocks, empty or malformed observations, or tool calls inconsistent with the recorded state transitions.
We additionally require a minimum exploration depth and at least one substantive observation, such as a parsed schema, computed statistic, cross-file correspondence, data-quality issue, or repository-level structural finding.
These criteria exclude traces that are well-formed but lack meaningful scientific evidence.

\noindent\textbf{Step-level hallucination detection.} For each retained trajectory, an LLM verifier checks two transitions at every step: whether action $a_t$ follows logically from the preceding reasoning and trajectory state, and whether the subsequent reasoning is grounded in observation $o_t$.
Each transition is labeled as supported, unsupported, or contradicted.
Branches containing unsupported or contradicted transitions are removed when the issue affects later actions, the evidence target, or the final answer.


\noindent\textbf{Uncertainty handling.} Ambiguous instances are withheld from automatic acceptance.
When ambiguity arises from processing failures, such as parsing errors, truncated outputs, or verifier instability, verification is repeated against the original execution trace.
Instances whose evidence remains intrinsically ambiguous are discarded.
This conservative protocol preserves the central supervision objective of \textit{SciDataSailor}: grounding final responses and intermediate reasoning in recoverable tool observations.

\subsection{Dataset Analysis}
\label{sec:dataset_analysis}

Figure~\ref{fig:data_analysis} in Appendix~\ref{app:data_statistics} summarizes answer length and interaction complexity for the benchmark and SFT corpus.
Figure~\ref{fig:benchmark_overview} details the benchmark, while Figure~\ref{fig:supp_sft_2k} in Appendix~\ref{app:sft-2k-overview} presents the composition of the SFT corpus.
To reduce distribution shift, we select raw datasets such that the training and evaluation splits remain aligned in answer length, ReAct-step budget, reasoning depth, and repository structure.
This alignment allows improvements learned from SFT trajectories to transfer to the benchmark without relying on a substantially easier or structurally different evaluation distribution.

\noindent\textbf{Answer-token length.} Answer-token length measures the compression and reporting burden of the final response.
Across both splits, QA answers are intentionally compact, while meta-information extraction requires longer evidence reports that summarize file organization, schemas, variables, statistics, and data-quality findings.
The SFT corpus is dominated by short answers but retains a long tail of 600--3K-token responses, providing supervision for both concise question answering and complete scientific data-analysis reports.
Across the benchmark, life, earth, and physical sciences account for 1,521, 1,160, and 1,642 tool calls and 4.01M, 3.27M, and 4.46M trajectory tokens, respectively.

\noindent\textbf{Tool-call count.} Tool-call count measures how many executable interactions an agent needs to complete a task.
Across the benchmark and SFT corpus, QA instances concentrate around a small number of tool calls, supporting stable and reproducible evaluation.
In contrast, meta-information extraction uses more tool calls, reflecting the long-horizon process. 

\noindent\textbf{Trace depth.} Trace depth captures the multi-hop reasoning horizon behind each answer.
QA trajectories are mostly shallow-to-moderate, whereas meta-information extraction exhibits deeper traces.
This is expected because meta-information tasks require agents to iteratively inspect, parse, verify, and synthesize observations rather than produce an answer from a single file or table.

\noindent\textbf{File-system nesting depth.} File-system nesting depth measures the structural complexity of scientific repositories.
The benchmark covers both flat repositories and deeply nested scientific data layouts, ensuring that evaluation does not collapse into a single directory-structure pattern.

%% file: components/4-Experiments.tex
\section{Experiments}
\label{sec:exp}

\subsection{Experiment Setup}
\label{subsec:exp_setup}

\input{components/table-main-results}
\input{components/table-qa-results}

\noindent\textbf{Models and agent scaffold.}
We evaluate all LLM backbones using a unified ReAct-style agent scaffold~\cite{yao2023react}, with Python execution implemented through a CodeAct-style interaction protocol~\cite{wang2024codeact}.
The proprietary models are GPT-5.4 \cite{gpt-5-4-thinking}, Gemini-3.1-Pro \cite{gemini-3-1-pro}, and Claude-Opus-4-7-Thinking \cite{opus-4-7}; the open-weight models are GLM-5.1 \cite{zeng2026glm}, Kimi-K2.6 \cite{kimi-2-6}, DeepSeek-V4-Pro and DeepSeek-V4-Flash \cite{xu2026deepseek}, GPT-OSS-120B and 20B \cite{agarwal2025gpt}, Qwen3-32B \cite{yang2025qwen3}, Qwen3.5-35B-A3B, Qwen3.5-27B, and Qwen3.5-9B \cite{qwen-3-5}.
For both benchmark tracks, we additionally evaluate Qwen3.5-9B-SFT, which is fine-tuned on our trajectory corpus.
Keeping the agent scaffold fixed isolates differences in planning, executable analysis, observation interpretation, and error recovery.

\noindent\textbf{Benchmark and metrics.}
\textit{SciDataSailor-Bench} contains complementary meta-information summarization \textit{SciDataSailor-Bench-Meta} and scientific question-answering \textit{SciDataSailor-Bench-QA}.
The former requires repository-level descriptions of organization, formats, schemas, metadata, provenance, and usage constraints; the latter requires agents to locate relevant files, execute analyses, and return data-grounded answers.
Meta-information responses are scored against trajectory-derived references using an LLM judge~\cite{zheng2023judging,qiao2026datamind,yang2026tooltree}, with prompts provided in Appendix~\ref{app:llm_judge_prompts}.
Scientific QA uses tolerance-aware matching for numerical answers and textual, categorical, and unit-consistency checks for non-numerical answers~\cite{dong2026agentic}.
One \emph{ReAct step} is a complete thought--code/action--observation cycle; its action may contain one or more parallel Python tool calls.
We report Pass@1, the success rate (SR) of producing a completed trajectory within the ReAct-step budget, and the average number of ReAct steps.

\noindent\textbf{Execution protocol.}
Agents operate in a controlled sandbox with Python as the only tool; web search, external retrieval, and auxiliary tools are disabled.
All models receive the same task contract, sandbox, and scoring procedure and are tested with maximum ReAct-step budgets of 12 and 24.
Serializer-specific XML, TIR, and function-calling adapters are documented in Appendix~\ref{app:prompt_protocols}.
We use a sampling temperature of 0.6 and top-$p=0.95$.

\input{components/table_ablation}

\subsection{Experimental Results}
\label{subsec:exp_results}
\noindent\textbf{Meta-information exploration.}
Table~\ref{tab:main_results} reveals a pronounced reliability and efficiency gap: with 12 ReAct steps, GPT-5.4 attains 65.22 Pass@1 and 98.07 SR in 6.03 average steps, whereas the strongest open-weight Pass@1 is 27.54 from GPT-OSS-120B.
Increasing the budget to 24 steps produces comparatively modest gains for proprietary models but substantially improves several open-weight systems: DeepSeek-V4-Pro, for example, rises from 16.43 to 52.66 Pass@1 and from 28.99 to 79.71 SR.
As visualized in Figure~\ref{fig:behavior_gap}, this sensitivity indicates that open-weight agents more often rely on extended trial-and-error exploration, whereas proprietary agents reach successful repository-level summaries more directly.

\noindent\textbf{Scientific question answering.}
The gap is considerably smaller on \textit{SciDataSailor-Bench-QA} as shown in Table~\ref{tab:qa_results}, where Claude-Opus-4-7-Thinking achieves the highest Pass@1 at both budgets (64.16 and 66.04), closely followed by DeepSeek-V4-Pro (61.43 and 65.53).
At 24 steps, Kimi-K2.6, GLM-5.1, and Qwen3.5-27B also exceed 63 Pass@1, showing that strong open-weight agents can approach proprietary performance when the task targets a specific answer rather than a repository-level synthesis.
The divergence between Pass@1 and SR for some models further shows that completing a trajectory does not guarantee a correct data-grounded answer.

\noindent\textbf{Effect of trajectory supervision.}
Fine-tuning Qwen3.5-9B on \textit{SciDataSailor-SFT-2K} consistently improves both benchmark tracks, increasing answer accuracy and trajectory completion while reducing the required ReAct steps (Tables~\ref{tab:main_results} and~\ref{tab:qa_results}).
The gains are most pronounced under constrained budgets, particularly for repository-level meta-information exploration, indicating that trajectory supervision improves both accuracy and exploration efficiency.

\subsection{Ablation Analysis}
\label{subsec:ablation}

We conduct a deterministic behavioral analysis across 207 paired tasks to characterize differences in execution efficiency, error resilience, and code-generation structure.
In Table~\ref{tab:behavior}, we compares GPT-5.4, DeepSeek-V4-Pro, Qwen3.5-35B-A3B, and the fine-tuned Qwen3.5-9B model under tool-call budgets of 12 and 24.

\noindent\textbf{Efficiency and turn-budget sensitivity.} Models split into two distinct profiles. GPT-5.4 and the SFT-trained 9B model complete tasks in approximately 5 tool calls regardless of turn budget, consuming 685K--759K total tokens, while DeepSeek-v4-pro and the Qwen3.5-35B are heavily constrained at T=12 -- 71\% and 46\% of their tasks hit the turn ceiling, and doubling to T=24 inflates their token consumption by 53\% and 34\%. The SFT-trained model matches the efficiency of GPT-5.4, a substantially larger proprietary model, while using 20\% fewer total tokens than the Qwen3.5-35B base at T=12 (759K vs. 950K). This demonstrates that SFT on high-quality trajectories teaches the model to plan and execute decisively: each tool call attempts to solve a large portion of the task in one shot, rather than relying on extended iterative exploration across many turns.

\noindent\textbf{Error resilience and failure isolation.} Per-call error rates are comparable across open models (5--12\%), but the Error-Free Trajectory Rate reveals that the SFT-trained 9B model produces error-free trajectories at a higher rate than the 35B models (53.\% vs. 45.9\% at T=12), despite having 3.7x fewer active parameters. When errors do occur, they remain isolated: the fine-tuned model exhibits the shortest Mean Error Chain Length among all models (1.07 vs. 1.33 for the 35B model), indicating that its errors are single-event failures rather than cascading sequences. The 35B model, by contrast, more frequently enters loops of 2--3 consecutive failing calls before recovery. These results suggest that the SFT dataset instills clean failure boundaries -- the model learns when to abandon a failing approach entirely rather than repeatedly patching it.

\noindent\textbf{Code density and structural complexity.} The most pronounced effect of SFT appears in code structure. The fine-tuned 9B model generates code blocks averaging 74.9 lines with 11.4 control-flow nodes and an AST depth of 12.3 -- nearly double the control-flow complexity (vs. 5.8--7.2 for other models) and 1.4--2.5x the code density. This reflects a "monolithic solver" strategy transferred from the curated trajectories: each call emits a comprehensive, self-contained program with rich branching and looping logic that addresses the full scope of a subtask in a single execution, rather than decomposing work into small sequential steps. This strategy trades iterative refinement for first-attempt completeness, a favorable tradeoff given that over half of all trajectories require no error recovery. By contrast, GPT-5.4 favors modular decomposition with shorter blocks (29.7 lines) but broader library diversity (6.7 unique imports/task vs. 4.2), while DeepSeek-v4-pro and the 35B base occupy an intermediate position with moderate code density and iterative debugging capability.

%% file: components/table-main-results.tex
\begin{table}[t]
\centering
\caption{\textbf{Performance comparison on \textit{SciDataSailor-Bench-Meta}} under ReAct-step budgets of 12 and 24. Pass@1 and success rate (SR) are percentages. Bold marks the best value in each column within each model group}
\label{tab:main_results}
\footnotesize
\setlength{\tabcolsep}{2.5pt}
\renewcommand{\arraystretch}{1.08}
\begin{tabularx}{\textwidth}{@{}>{\raggedright\arraybackslash}p{0.22\textwidth}YYYYYY@{}}
\toprule
\multirow{2}{*}{\textbf{Backbone}}
& \multicolumn{3}{c}{\textbf{Max. ReAct Steps: 12}}
& \multicolumn{3}{c}{\textbf{Max. ReAct Steps: 24}} \\
\cmidrule(lr){2-4} \cmidrule(lr){5-7}
& \textbf{Pass@1} & \textbf{SR} & \textbf{Avg. Steps}
& \textbf{Pass@1} & \textbf{SR} & \textbf{Avg. Steps} \\
\midrule

\multicolumn{7}{c}{\cellcolor{gray!20}\textbf{Proprietary Models}} \\
\midrule
GPT-5.4                                    & \textbf{65.22} & \textbf{98.07} & \textbf{6.03} & \textbf{69.57} & \textbf{100.00} & \textbf{6.10} \\
Gemini-3.1-Pro                             & 43.48 & 86.96 & 6.96 & 51.21 & 99.52 & 7.32 \\
Claude-Opus-4-7-Thinking                   & 51.69 & 77.29 & 8.51 & 66.67 & 99.03 & 9.70 \\
\midrule

\multicolumn{7}{c}{\cellcolor{gray!20}\textbf{Open-weight Models ($>30$B)}} \\
\midrule
GLM-5.1                                    & 4.35 & 7.73 & 11.78 & 19.32 & 35.27 & 21.26 \\
Kimi-K2.6                                  & 14.49 & 29.95 & 10.70 & 43.48 & 76.33 & 15.68 \\
DeepSeek-V4-Pro                            & 16.43 & 28.99 & 11.05 & \textbf{52.66} & 79.71 & 15.49 \\
DeepSeek-V4-Flash                          & 11.11 & 24.15 & 11.41 & 46.86 & 78.26 & 17.86 \\
GPT-OSS-120B                               & \textbf{27.54} & \textbf{66.67} & \textbf{9.39} & 34.30 & \textbf{97.58} & \textbf{10.88} \\
\midrule
\multicolumn{7}{c}{\cellcolor{gray!20}\textbf{Open-weight Models ($<30$B)}} \\
\midrule
Qwen3.5-27B                                & 14.01 & 35.75 & 11.18 & 41.06 & 95.17 & 14.76 \\
Qwen3.5-35B-A3B                            & 18.84 & 47.34 & 10.14 & \textbf{45.89} & 91.30 & 12.60 \\
Qwen3-32B                                  & 14.98 & 86.48 & \textbf{5.69} & 13.53 & 93.72 & \textbf{5.67} \\
GPT-OSS-20B                                & 9.66 & 44.93 & 10.49 & 16.42 & 85.02 & 14.65 \\
Qwen3.5-9B                                 & 14.01 & 49.76 & 10.21 & 22.71 & 70.05 & 9.42 \\
\rowcolor{cyan!15}
Qwen3.5-9B-SFT                             & \textbf{28.99} {\scriptsize\textcolor{green!50!black}{$\uparrow14.98$}} & \textbf{96.14} {\scriptsize\textcolor{green!50!black}{$\uparrow46.38$}} & 6.24 {\scriptsize\textcolor{green!50!black}{$\downarrow3.97$}}
                                            & 24.64 {\scriptsize\textcolor{green!50!black}{$\uparrow1.93$}} & \textbf{96.14} {\scriptsize\textcolor{green!50!black}{$\uparrow26.09$}} & 5.92 {\scriptsize\textcolor{green!50!black}{$\downarrow3.50$}} \\
\bottomrule
\end{tabularx}
\end{table}

%% file: components/table-qa-results.tex
\begin{table}[t]
\centering
\caption{\textbf{Performance comparison on \textit{SciDataSailor-Bench-QA}} under ReAct-step budgets of 12 and 24. Pass@1 and success rate (SR) are percentages. Bold marks the best value in each column within each model group (higher for Pass@1 and SR; lower for Avg. Steps); arrows report SFT changes from Qwen3.5-9B.}
\label{tab:qa_results}
\footnotesize
\setlength{\tabcolsep}{2.5pt}
\renewcommand{\arraystretch}{1.08}
\begin{tabularx}{\textwidth}{@{}>{\raggedright\arraybackslash}p{0.22\textwidth}YYYYYY@{}}
\toprule
\multirow{2}{*}{\textbf{Backbone}}
& \multicolumn{3}{c}{\textbf{Max. ReAct Steps: 12}}
& \multicolumn{3}{c}{\textbf{Max. ReAct Steps: 24}} \\
\cmidrule(lr){2-4} \cmidrule(lr){5-7}
& \textbf{Pass@1} & \textbf{SR} & \textbf{Avg. Steps}
& \textbf{Pass@1} & \textbf{SR} & \textbf{Avg. Steps} \\
\midrule

\multicolumn{7}{c}{\cellcolor{gray!20}\textbf{Proprietary Models}} \\
\midrule
GPT-5.4                                    & 61.95 & \textbf{97.78} & \textbf{5.16} & 62.97 & \textbf{99.66} & \textbf{5.27} \\
Gemini-3.1-Pro                             & 60.58 & 86.01 & 6.75 & 60.92 & 92.66 & 7.82 \\
Claude-Opus-4-7-Thinking                   & \textbf{64.16} & 94.71 & 5.34 & \textbf{66.04} & 99.15 & 5.55 \\
\midrule

\multicolumn{7}{c}{\cellcolor{gray!20}\textbf{Open-weight Models ($>30$B)}} \\
\midrule
GLM-5.1                                    & 58.02 & 79.52 & 7.94 & 63.82 & 92.66 & 9.48 \\
Kimi-K2.6                                  & 56.66 & 62.46 & 9.01 & 64.85 & 87.54 & 11.49 \\
DeepSeek-V4-Pro                            & \textbf{61.43} & \textbf{81.74} & 7.58 & \textbf{65.53} & 94.20 & 8.37 \\
DeepSeek-V4-Flash                          & 56.66 & 74.40 & 8.64 & 61.77 & 91.81 & 10.46 \\
GPT-OSS-120B                               & 40.61 & 77.13 & \textbf{7.48} & 25.77 & \textbf{98.63} & \textbf{5.98} \\
\midrule
\multicolumn{7}{c}{\cellcolor{gray!20}\textbf{Open-weight Models ($<30$B)}} \\
\midrule
Qwen3.5-27B                                & \textbf{56.66} & 77.82 & 8.24 & \textbf{63.31} & \textbf{95.56} & 9.63 \\
Qwen3.5-35B-A3B                            & 52.05 & 74.91 & 8.45 & 59.04 & 92.83 & 10.14 \\
Qwen3-32B                                  & 7.34 & 82.59 & \textbf{5.54} & 7.85 & 84.30 & \textbf{6.72} \\
GPT-OSS-20B                                & 20.48 & 70.14 & 7.88 & 31.23 & 89.93 & 10.85 \\
Qwen3.5-9B                                 & 42.32 & 71.50 & 7.91 & 48.12 & 85.49 & 9.45 \\
\rowcolor{cyan!15}
Qwen3.5-9B-SFT                             & 48.81 {\scriptsize\textcolor{green!50!black}{$\uparrow6.49$}} & \textbf{82.76} {\scriptsize\textcolor{green!50!black}{$\uparrow11.26$}} & 7.27 {\scriptsize\textcolor{green!50!black}{$\downarrow0.64$}}
                                            & 51.19 {\scriptsize\textcolor{green!50!black}{$\uparrow3.07$}} & 93.17 {\scriptsize\textcolor{green!50!black}{$\uparrow7.68$}} & 7.68 {\scriptsize\textcolor{green!50!black}{$\downarrow1.77$}} \\
\bottomrule
\end{tabularx}
\end{table}

%% file: components/table_ablation.tex
\begin{table}[t]
\centering
\caption{\textbf{Behavioral comparison of agent trajectories} across four representative models under maximum ReAct-step budgets of 12 and 24. Each cell reports the 12-step\,/\,24-step value. Bold indicates the preferred value for efficiency and error metrics and the largest value for descriptive code-structure metrics.}
\label{tab:behavior}
\small
\setlength{\tabcolsep}{4pt}
\renewcommand{\arraystretch}{1.08}
\begin{tabular*}{\textwidth}{@{\extracolsep{\fill}}lcccc@{}}
\toprule
\textbf{Metric}
& \textbf{GPT-5.4}
& \textbf{DeepSeek-V4-Pro}
& \textbf{Qwen3.5-35B-A3B}
& \textbf{Qwen3.5-9B-SFT} \\
\midrule

\multicolumn{5}{c}{\cellcolor{gray!20}\textbf{Efficiency}} \\
\midrule
Avg. ReAct Steps\,/\,Task 
  & 5.76\,/\,5.76
  & 10.92\,/\,15.14
  & 9.67\,/\,11.69
  & \textbf{5.28}\,/\,\textbf{4.97} \\
Total Tokens ($\times 10^3$) 
  & \textbf{685}\,/\,\textbf{695}
  & 1{,}024\,/\,1{,}565
  & 950\,/\,1{,}274
  & 759\,/\,700 \\
\midrule

\multicolumn{5}{c}{\cellcolor{gray!20}\textbf{Error and Recovery}} \\
\midrule
Code Error Rate (\%) 
  & 7.22\,/\,6.47
  & \textbf{5.14}\,/\,\textbf{5.79}
  & 11.37\,/\,12.24
  & 10.48\,/\,11.45 \\
Error-Free Trajectory Rate (\%)
  & \textbf{69.08}\,/\,\textbf{72.46}
  & 63.29\,/\,46.86
  & 45.89\,/\,41.06
  & 53.62\,/\,54.11 \\
Mean Error Chain Length
  & 1.13\,/\,1.15
  & 1.14\,/\,1.15
  & 1.33\,/\,1.39
  & \textbf{1.07}\,/\,\textbf{1.13} \\
\midrule

\multicolumn{5}{c}{\cellcolor{gray!20}\textbf{Code Structure Sophistication}} \\
\midrule
Code Nesting Depth
  & 11.29\,/\,11.27
  & 10.71\,/\,10.66
  & 10.65\,/\,10.68
  & \textbf{12.34}\,/\,\textbf{12.45} \\
Branches \& Loops
  & 5.84\,/\,5.94
  & 6.46\,/\,6.65
  & 6.63\,/\,7.20
  & \textbf{11.39}\,/\,\textbf{10.98} \\
Lines of Code
  & 29.66\,/\,30.32
  & 41.67\,/\,43.99
  & 52.81\,/\,58.21
  & \textbf{74.86}\,/\,\textbf{72.89} \\
\bottomrule
\end{tabular*}
\end{table}

%% file: components/5-Conclusion.tex
\section{Conclusion and Discussion}
\label{sec:conclusion}
Scientific data exploration presents a distinct challenge for LLM agents because evidence is distributed across heterogeneous files, implicit schemas, metadata, and derived artifacts that cannot be reliably reconstructed from text retrieval or parametric knowledge alone.
This setting therefore provides a meaningful task paradigm for developing agents that must discover, execute, verify, and integrate evidence in open-ended environments.
Our evaluation reveals a pronounced behavioral gap: proprietary models explore scientific repositories more consistently and efficiently, whereas open-weight models often depend on additional trial-and-error interactions.
This gap suggests that the central bottleneck is not isolated tool use, but maintaining a coherent evidence-acquisition policy across uncertain and interdependent repository states.
Although supervised fine-tuning on verified trajectories transfers effective decision patterns and substantially improves efficiency, token-level imitation provides limited support for recovering from unforeseen execution failures.
Combining trajectory supervision with verifier-guided reinforcement learning is a promising direction for strengthening these capabilities and enabling adaptive self-correction.

%% file: components/X-appendix.tex
\lstdefinestyle{seedcontent}{
  basicstyle=\ttfamily\scriptsize,
  breaklines=true,
  breakatwhitespace=false,
  columns=fullflexible,
  keepspaces=true,
  frame=single,
  xleftmargin=1em,
  xrightmargin=1em,
  showstringspaces=false
}

\lstdefinestyle{judgeprompt}{
  basicstyle=\ttfamily\scriptsize,
  breaklines=true,
  breakatwhitespace=false,
  breakautoindent=false,
  breakindent=0pt,
  columns=fullflexible,
  keepspaces=true,
  frame=single,
  xleftmargin=1em,
  xrightmargin=1em,
  showstringspaces=false
}

\definecolor{appendixPromptBg}{RGB}{247,247,247}
\definecolor{appendixTraceBg}{RGB}{250,250,250}
\definecolor{appendixRule}{RGB}{190,190,190}
\lstdefinestyle{appendixprompt}{
  basicstyle=\ttfamily\scriptsize,
  backgroundcolor=\color{appendixPromptBg},
  frame=single,
  rulecolor=\color{appendixRule},
  breaklines=true,
  breakatwhitespace=false,
  columns=fullflexible,
  keepspaces=true,
  showstringspaces=false,
  tabsize=2,
  aboveskip=0.8em,
  belowskip=0.9em,
  captionpos=t
}
\lstdefinestyle{appendixtrace}{
  style=appendixprompt,
  backgroundcolor=\color{appendixTraceBg},
  numbers=left,
  numberstyle=\tiny\color{gray},
  numbersep=6pt,
  xleftmargin=1.4em,
  framexleftmargin=1.1em
}

\input{components/appendix/dataset-training.tex}
\input{components/appendix/prompt-evaluation.tex}
\input{components/appendix/budget-case-study.tex}

%% file: components/appendix/dataset-training.tex
\section{Dataset Construction and Training Details}
\label{app:technical_appendices}

This section documents the benchmark seed taxonomy, training-corpus composition,
dataset statistics, and supervised fine-tuning configuration needed to reproduce
the reported data construction and training setup.

\subsection{Seed Difficulty Stratification}
\label{app:seed_levels}
We stratify scientific data exploration seeds into four executable difficulty levels:
\begin{itemize}
    \item \textbf{L0}
    focuses on building a dataset map with little relational reasoning.
    The corresponding seed intents are:
\begin{lstlisting}[style=seedcontent]
[
{"content": "** Dataset structure exploration **: Explore the dataset hierarchy by enumerating files and subdirectories, identifying file formats, recording file sizes, and summarizing how raw data, metadata, annotations, intermediate products, and derived outputs are organized. The goal is to produce a high-level map of the dataset structure that makes the storage logic and processing stages understandable before deeper analysis begins."},
{"content": "** File format and content profiling **: Inspect the major file types in the dataset and characterize their internal content, such as tabular fields, image dimensions, signal channels, matrix shapes, text schemas, or serialization formats. This task aims to clarify what kinds of information each file type contains, how those files are intended to be interpreted, and whether their internal organization is consistent across the dataset."},
{"content": "** Modality, variable, and annotation inventory **: Catalog the data modalities, variables, annotation fields, labels, measurement types, and auxiliary metadata present in the dataset, and summarize their semantic roles in the overall experimental design. The goal is to establish a clear inventory of what information is actually available, which variables are central versus peripheral, and how well the annotations support different forms of scientific analysis."}
]
\end{lstlisting}

    \item \textbf{L1}
    introduces field semantics, basic statistics, and lightweight evidence-based checks.
    The corresponding seed intents are:
\begin{lstlisting}[style=seedcontent]
[
{"content": "** Numerical statistics and distribution analysis **: Compute descriptive statistics for numerical data files, compare distributions across relevant samples, subsets, batches, or experimental conditions, and identify notable patterns such as outliers, missingness, skewed distributions, abnormal ranges, or unexpected shifts. The objective is to build an initial quantitative understanding of the dataset and detect statistical irregularities that may influence downstream analysis."},
{"content": "** Data quality assessment **: Evaluate data quality by checking completeness, duplicate records or files, schema consistency, identifier integrity, value validity, formatting irregularities, and other issues that could compromise analysis reliability. This task is intended to identify practical problems in the dataset and provide an evidence-based summary of quality risks that should be addressed before modeling or scientific interpretation."}
]
\end{lstlisting}

    \item \textbf{L2}
    requires cross-file alignment, entity mapping, or multi-source consistency reasoning.
    The corresponding seed intents are:
\begin{lstlisting}[style=seedcontent]
[
{"content": "** Cross-file metadata and measurement integration **: Cross-reference multiple file types to construct a unified view of sample metadata, experimental measurements, annotations, and derived values, while preserving identifier alignment and provenance. This task focuses on determining how separate files complement one another and whether they can be reliably linked into an integrated dataset representation."},
{"content": "** Entity and sample relationship mapping **: Identify the relationships among key dataset entities, such as samples, subjects, sessions, experiments, runs, time points, regions, or modalities, and reconstruct how these entities are connected across files. The purpose is to reveal the relational structure of the dataset so that downstream analysis can correctly interpret repeated measures, hierarchical grouping, longitudinal structure, or many-to-one mappings."}
]
\end{lstlisting}

    \item \textbf{L3}
    targets scientific usability, robustness, and alternative-explanation checks.
    The corresponding seed intent is:
\begin{lstlisting}[style=seedcontent]
[
{"content": "** Scientific usability and downstream question assessment **: Assess which scientific questions the dataset can realistically support by examining sample diversity, annotation richness, feature coverage, temporal or spatial scope, modality availability, and the alignment between available variables and plausible downstream tasks. This task moves from data inspection to scientific utility evaluation, aiming to determine the analytical potential and practical limitations of the dataset for future research use."}
]
\end{lstlisting}
\end{itemize}

\subsection{Overview of SciDataSailor-SFT-2K}
\label{app:sft-2k-overview}

\begin{figure}[t]
  \centering
  \includegraphics[width=\textwidth]{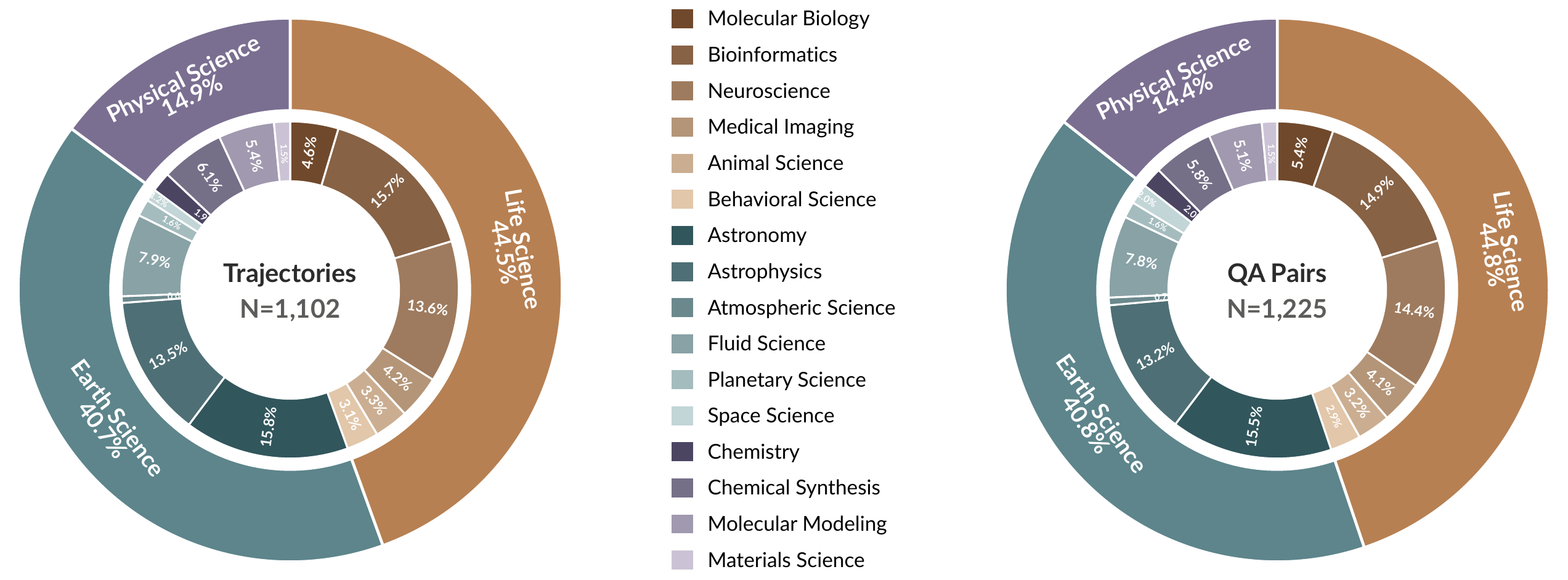}
  \caption{
    \textbf{Distribution of the \textit{SciDataSailor-SFT-2K} dataset.}
    \textit{SciDataSailor-SFT-2K} contains 1,102 meta-information exploration trajectories and 1,225 dataset-targeted QA pairs constructed from heterogeneous datasets across the life, earth, and physical sciences.
    The two circular taxonomy plots summarize the distribution of trajectory and QA samples over source datasets, showing broad coverage across scientific domains rather than concentration on a small set of tasks.
    The bottom bars group the samples by high-level scientific discipline, highlighting the dataset's cross-domain composition and suitability for supervised fine-tuning of scientific data exploration models.}
  \Description{Two circular taxonomy plots and grouped bars summarize the distribution of Meta trajectories and QA pairs across source datasets and scientific disciplines.}
  \label{fig:supp_sft_2k}
\end{figure}

Figure~\ref{fig:supp_sft_2k} summarizes the composition of \textit{SciDataSailor-SFT-2K}.
The dataset contains both tool-use trajectories and evidence-grounded QA pairs, exposing models to complementary supervision signals: long-horizon exploration behavior and compact answer synthesis.
Its samples are distributed across life sciences, earth sciences, and physical sciences, with meta-information trajectories and QA instances drawn from a broad set of source datasets rather than a single dominant repository.
This cross-domain coverage encourages models to learn reusable scientific data exploration skills, including file-system navigation, schema inspection, statistical probing, and evidence integration across heterogeneous data formats.

\subsection{Dataset Statistics}
\label{app:data_statistics}

\begin{figure}[t]
  \centering
  \includegraphics[width=\textwidth]{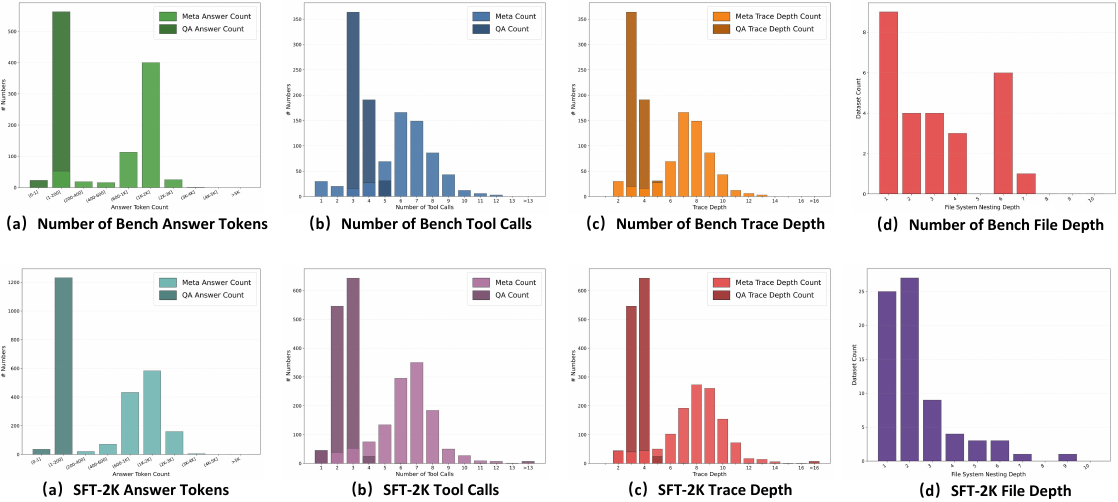}
  \caption{\textbf{Data statistics of \textit{SciDataSailor-Bench} and \textit{SciDataSailor-SFT-2K}.} The top row summarizes \textit{SciDataSailor-Bench} and the bottom row summarizes \textit{SciDataSailor-SFT-2K} in terms of answer-token length, tool-call count, trace depth, and file-system nesting depth when available.}
  \Description{Eight statistical plots compare answer length, tool-call count, trace depth, and file-system nesting depth for the benchmark and supervised fine-tuning corpus.}
  \label{fig:data_analysis}
\end{figure}

Figure~\ref{fig:data_analysis} compares the answer length and interaction complexity of \textit{SciDataSailor-Bench} and \textit{SciDataSailor-SFT-2K}.

\subsection{Supervised Fine-Tuning Details}
\label{app:sft_setup}

During supervised fine-tuning, we follow the ARPO~\cite{dong2026agentic} training protocol and train Qwen3.5-9B on \textit{SciDataSailor-SFT-2K} using the LLaMA-Factory framework~\cite{zheng2024llamafactory}.
Training is conducted on 4 NVIDIA H200 GPUs with a learning rate of $7\times 10^{-6}$.
We employ DeepSpeed ZeRO-2~\cite{rasley2020deepspeed} and FlashAttention-2~\cite{dao2023flashattention2} for memory and attention optimization.
The global batch size is set to 8 with a weight decay of 0.1, and the model is trained for 20 epochs.
We use BF16 mixed precision with a maximum input length of 4096 tokens.

%% file: components/appendix/prompt-evaluation.tex
\section{Prompt and Evaluation Protocols}
\label{app:prompt_evaluation_protocols}

This section separates the model-visible trajectory prompts from the prompts
used to construct references and score final answers. This distinction makes
the data-generation, evaluation-adapter, and judging stages independently
auditable.

\subsection{Agent Prompt Protocols}
\label{app:prompt_protocols}
\label{app:prompt_data_synthesis}

\paragraph{Meta trajectory system prompt.}
The original Meta trajectories use an XML-serialized ReAct-style protocol.
The prompt defines the accessible data boundary, makes Python execution the
only observation channel, and requires a single final answer after iterative
\texttt{think--python--result} blocks.  Listing~\ref{lst:meta-system-prompt}
reproduces the exact \texttt{system} field stored in the Meta SFT records,
with no dataset-specific content interpolated.

\begin{lstlisting}[style=appendixprompt,caption={Meta-task system prompt stored in the original trajectory data.},label={lst:meta-system-prompt}]
You are a scientific data exploration assistant that solves a question step by step with the help of a python interpreter tool. You first think about the reasoning process in your mind and then provide the final answer. During thinking, you can invoke the python interpreter to traverse folders, read files and compute facts about the dataset.

The reasoning process and final answer are enclosed within <think> </think> and <answer> </answer> tags respectively; python code and its result are enclosed within <python> </python> and <result> </result> tags respectively. After receiving a python result, you continue your reasoning from a new <think>. For example: <think> reasoning </think> <python> python code here </python> <result> tool output here </result> <think> more reasoning </think> <python> python code here </python> <result> tool output here </result> <think> final reasoning </think> <answer> The final answer is \[ \boxed{answer here} \] </answer>. In the <answer> block, the final exact answer MUST be enclosed within \boxed{} in latex format.
\end{lstlisting}

\paragraph{QA trajectory system prompt.}
The QA source repository stores Tool-Integrated Reasoning (TIR) trajectories.
Its authored system preamble is shorter because action/result serialization is
handled by the sampler and conversion code.  It nevertheless imposes the same
core contract: inspect the local dataset with Python, ground claims in observed
files, and return a self-contained answer.  Listing~\ref{lst:qa-system-prompt}
is the exact preamble imported by the QA-to-TIR conversion pipeline.

\begin{lstlisting}[style=appendixprompt,caption={QA-task system prompt used by the source TIR trajectory pipeline.},label={lst:qa-system-prompt}]
You are a scientific data exploration assistant that solves a question step by step with the help of a python interpreter tool. You first think about the reasoning process in your mind and then provide the final answer. During thinking, you can invoke the python interpreter to traverse folders, read files and compute facts about the dataset.

The reasoning process and final answer are enclosed within <think> </think> and <answer> </answer> tags respectively; python code and its result are enclosed within <python> </python> and <result> </result> tags respectively. After receiving a python result, you continue your reasoning from a new <think>. For example: <think> reasoning </think> <python> python code here </python> <result> tool output here </result> <think> more reasoning </think> <python> python code here </python> <result> tool output here </result> <think> final reasoning </think> <answer> The final answer is \[ \boxed{answer here} \] </answer>. In the <answer> block, the final exact answer MUST be enclosed within \boxed{}.

Output-length discipline:
- Only inspect files/dirs under the provided dataset root path.
- Keep outputs concise: report key findings, not full file dumps.
- Never print huge recursive listings or thousands of matches/rows.
- For large outputs, print a small sample + aggregate counts/statistics.
\end{lstlisting}

\paragraph{Function-calling evaluation adapter.}
The GPT-5.4 and DeepSeek-V4-Pro runs in
Section~\ref{app:budget_case} use structured API tool calls rather
than literal XML tags.  Listing~\ref{lst:adapter-system-prompt} is taken from
the saved run artifact itself.  We replace only the machine-specific absolute
path by \texttt{\{DATASET\_ROOT\}}.  The selected artifacts do not contain the
optional ReAct-step-budget hint present in newer source revisions, so that hint is
not retroactively added here.

\begin{lstlisting}[style=appendixprompt,caption={Function-calling system prompt captured in the evaluated run.},label={lst:adapter-system-prompt}]
You are a scientific data exploration assistant that solves a question step by step with the help of a Python interpreter tool.
You can call the function tool named `python_interpreter` when you need to traverse folders, read files, or compute facts about the dataset.

Use the OpenAI-compatible tool-calling API for tool use. Do not write manual XML tool calls, `<tool_call>` blocks, `<arg_key>`, or `<arg_value>` in message content.
When you need computation or file inspection, call `python_interpreter` with a JSON argument object containing one string field: `code`.
After receiving tool output, continue reasoning and call the tool again if more evidence is needed.

Output-length discipline:
- Only inspect files/dirs under the provided dataset root path.
- Keep outputs concise: report key findings, not full file dumps.
- Never print huge recursive listings or thousands of matches/rows.
- For large outputs, print a small sample plus aggregate counts/statistics.

Dataset-root discipline:
- The current dataset root path is: {DATASET_ROOT}
- Restrict file traversal and inspection to this root unless the user explicitly changes it.
- Prefer using this exact path in Python code rather than reconstructing it manually.

Necessary rules:
- Use only the `python_interpreter` tool for executable actions.
- Do not use web search, shell commands, CLI commands, notebook magics, or external tools.
- End with a final answer when the task is complete.
\end{lstlisting}

The adapter exposes one tool named \texttt{python\_interpreter}, which accepts
one string argument named \texttt{code}. Its state persists across calls,
whereas wall-clock time is bounded per call. One recorded ReAct step comprises
the model's thought, its code/action, and the resulting observation; the action
may issue multiple parallel calls. In
the selected Meta runs, the user message is one of the broad exploration
queries below.  QA is run through the legacy evaluator, where a
dataset-specific factual question is placed in a user envelope that names the
dataset root and asks the agent to use \texttt{final\_answer}.  Thus both task
families rely on local Python inspection, but they should not be described as
sharing an identical serializer.

\subsection{Reference Construction and LLM-as-a-Judge}
\label{app:llm_judge_prompts}

\subsubsection{Meta-Information Reference}
\label{app:meta_reference_synthesis_prompt}

For the meta-information track, we first synthesize multiple trajectory-derived candidate answers
into a single reference answer. The synthesis prompt is:

\begin{lstlisting}[style=judgeprompt]
[System]
You synthesize high-quality reference answers for scientific dataset meta-information tasks.

You will receive several candidate final answers produced from the same seed task and the same dataset. Your goal is to merge them into one comprehensive reference answer for later evaluation.

Preserve concrete and checkable information whenever it is present: dataset purpose and scientific domain; top-level directory and file organization; major file formats and storage modalities; schema fields, column names, array keys, metadata fields, and dtypes; row counts, sample counts, file sizes, split sizes, and numeric ranges; provenance, license, version, benchmark split, or configuration notes; known caveats, missing metadata, failed inspections, and uncertainty; intended storage logic, processing logic, and scientific usage constraints.

Do not over-compress the answer. Prefer a detailed but readable reference over a short abstract. Remove only duplicated wording, low-level tool traces, and claims that are clearly unsupported by all candidate answers.

When candidate answers conflict, do not invent a new fact. Prefer the fact that is supported by more candidates or by more specific evidence. If the conflict is material, mention the uncertainty briefly, for example by reporting a range or noting that different candidate answers disagree.

Do not add outside knowledge, web knowledge, or assumptions that are not supported by the candidate answers. Return only the synthesized reference answer, without JSON, bullet labels such as "analysis", or extra commentary.

[User]
Dataset category: {category}
Dataset name: {dataset}

Seed task:
{seed_task}

Candidate final answers to synthesize:

[Answer 1]
{candidate_answer_1}

[Answer 2]
{candidate_answer_2}

...
\end{lstlisting}

\subsubsection{LLM-as-a-Judge Scoring}
\label{app:judge_scoring_prompt}

The final answer is scored by comparing the candidate response with the reference answer. The judge
returns a binary score and a concise explanation in strict JSON format.

\begin{lstlisting}[style=judgeprompt]
[System]
You are a strict but fair evaluator for scientific dataset exploration tasks. Your job is to compare a candidate answer against a reference answer for the same question or seed task.

Judge semantic correctness, not surface form. Equivalent wording, ordering, and formatting are acceptable. Extra details are acceptable only when they do not contradict the reference. Penalize answers that contain major factual errors, unsupported numerical claims, wrong units, wrong dataset objects, or omissions of central reference facts.

For numerical answers, allow scientifically reasonable rounding, equivalent units, and formatting differences. Do not require exact string matching when the quantity is clearly the same. However, mark the answer incorrect if the numerical value, sign, scale, unit, or target entity is materially wrong.

For meta-information summaries, focus on whether the answer captures the core dataset-level evidence: organization, file formats, schemas, metadata fields, key counts/sizes, and storage or processing logic. Do not require every minor detail, but central facts in the reference must not be omitted or contradicted.

Return strict JSON only. Do not include markdown, explanations outside the JSON object, chain-of-thought, or additional keys.

[User]
Task type: {task_type}
Rubric: {rubric}

Question or task:
{question}

Reference answer:
{reference}

Candidate answer:
{prediction}

Return JSON exactly in this schema:
{"score": 0 or 1, "reason": "short reason"}.
Keep reason under 60 words. Do not output any extra text.
\end{lstlisting}

\subsubsection{Task-Specific Rubrics}
\label{app:judge_rubrics}

The \texttt{\{rubric\}} field is instantiated with one of the following task-specific criteria.

\noindent\textbf{Meta-information summary rubric.}

\begin{lstlisting}[style=judgeprompt]
Score 1 if the candidate captures the core metadata summary in the reference. The answer should cover the main dataset organization, major file formats, important schema or metadata fields, key counts or sizes when present, and the intended storage or processing logic.

Equivalent wording, reordered sections, and additional correct details are acceptable. The candidate does not need to reproduce every minor field or every number if the main metadata summary is faithful.

Score 0 if the candidate makes major factual errors, contradicts important reference facts, hallucinates unsupported counts or file structures, uses the wrong dataset, omits central schema/format information, or misses the main scientific usage constraints described in the reference.
\end{lstlisting}

\vspace{0.8em}

\noindent\textbf{Scientific dataset QA rubric.}

\begin{lstlisting}[style=judgeprompt]
Score 1 if the candidate answers the scientific dataset QA question with the same factual or numerical answer as the reference.

Equivalent units, rounding, significant figures, and wording are acceptable when scientifically consistent. If the reference answer contains multiple required values, all central values must be present and matched.

Score 0 if the candidate gives the wrong value, wrong unit, wrong entity, wrong comparison target, or a vague answer that does not resolve the question. Also score 0 when the answer is unsupported by the dataset evidence or contradicts the reference.
\end{lstlisting}

%% file: components/appendix/budget-case-study.tex
\clearpage
\section{Budget-Sensitive Trajectory Case Study}
\label{app:budget_case}

\paragraph{Selection rule.}
We joined the three judge files by sample UID and selected records satisfying
\(s_{\text{GPT-5.4},12}=1\),
\(s_{\text{DeepSeek},12}=0\), and
\(s_{\text{DeepSeek},24}=1\).  There are 62 such records; in 61 of their saved
12-ReAct-step inference records, DeepSeek explicitly terminates with
\texttt{max\_turns\_reached}.  To limit presentation-length selection bias, we
report the member of that 61-record subset with the shortest combined stored
transcript. For anonymous review, its UID is reported as
\texttt{RUN\_UID} and its machine-specific dataset path as
\texttt{DATASET\_ROOT}. The 12- and 24-step
DeepSeek traces are independent runs under different budgets, not a single
trajectory resumed at ReAct Step 13. Table~\ref{tab:appendix-guacamol-outcomes}
summarizes the three outcomes.

\begin{table}[!ht]
  \centering
  \scriptsize
  \caption{Outcome of the three runs for the selected sample. ``ReAct steps''
  counts complete thought--code/action--observation cycles; one action may
  contain multiple parallel tool calls.}
  \label{tab:appendix-guacamol-outcomes}
  \begin{tabular}{@{}lrrlc@{}}
    \toprule
    Model & Budget & ReAct steps & Stop reason & Judge score \\
    \midrule
    GPT-5.4          & 12 &  4 & \texttt{final\_response}     & 1 \\
    DeepSeek-V4-Pro  & 12 & 12 & \texttt{max\_turns\_reached} & 0 \\
    DeepSeek-V4-Pro  & 24 & 14 & \texttt{final\_response}     & 1 \\
    \bottomrule
  \end{tabular}
\end{table}

\paragraph{Task and reference answer.}
The user asks for the eighth seed, a scientific-usability assessment.  The
reference answer is shown in Listing~\ref{lst:guacamol-reference} before the
model traces so that the case can be read as a grounded comparison rather than
judged from fluency alone.

\begin{lstlisting}[style=appendixprompt,caption={Reference answer for the selected GuacaMol scientific-usability sample.},label={lst:guacamol-reference}]
GuacaMol supports generative modeling, distribution learning, and drug-like chemical space exploration, comprising 1.59 million unique molecules within the 200--500 Da weight range. While the dataset enables benchmarking for molecular generation and property prediction via computed scores, it lacks experimental property annotations and 3D structural data, limiting its direct applicability to experimental QSAR studies or structure-based drug discovery without supplementary data.
\end{lstlisting}

\paragraph{Trace rendering.}
Listings~\ref{lst:gpt54-trace}, \ref{lst:ds12-trace}, and
\ref{lst:ds24-trace} preserve every saved user message, available reasoning
field, tool call, tool observation, and final answer. We add only role/step
delimiters and the run-termination marker.
Typographic Unicode symbols are mapped to ASCII equivalents for LaTeX
portability; no substantive text or trajectory step is elided.  In particular,
the long tabular/statistical outputs remain part of the record.

\input{components/appendix/traces/gpt54-12.tex}
\input{components/appendix/traces/deepseek-v4-pro-12.tex}
\input{components/appendix/traces/deepseek-v4-pro-24.tex}

\subsection{What Changes When the Budget Doubles?}
\label{app:budget_analysis}

\begin{sloppypar}
GPT-5.4 first inventories the directory and then inspects the README and both
CSV files with parallel calls. By its fourth ReAct step it has computed schema,
missingness, uniqueness, length summaries, character sets, token frequencies,
and sample-level proxies.  Its answer is not merely a file listing: it connects
the observations to practical uses (model training and constrained generation)
and names unsupported analyses, which matches the reference answer's emphasis
on benchmark readiness and limitations.

The 12-step DeepSeek run spends its entire budget building increasingly rich
diagnostics.  It successfully verifies the train/validation split, SMILES
validity, duplicates, molecular-weight proxies, scaffold proxies, and token
coverage.  Its twelfth action, however, is another computation rather than a
submission, so the evaluator has no final answer to score.  This is a useful
failure mode: the evidence is largely present in the context, but the policy
does not reserve a ReAct step for synthesis.

With a 24-step allowance, DeepSeek follows a similar exploratory path but is
able to continue through stereochemistry, ring and heteroatom statistics,
property space, distribution shift, and diversity checks.  It submits at
Step 14 and converts
those observations into an explicit usability judgment with strengths,
limitations, and recommended validation steps.  The comparison therefore
supports a specific interpretation of the budget effect: additional steps help
not only by exposing more facts, but by creating room to stop exploring and
compose a judgeable answer.  Because the two DeepSeek runs are independent,
this case is evidence of budget sensitivity rather than proof that Step 13 alone
causally repairs the failed trajectory.
\end{sloppypar}

%% file: components/appendix/traces/gpt54-12.tex
\begin{lstlisting}[style=appendixtrace,caption={Complete GPT-5.4 trace with a 12-step budget (judge score: 1).},label={lst:gpt54-trace}]
RUN UID: {RUN_UID}
RECORDED STEPS: 4
STOP REASON: final_response

===== USER =====
** Scientific usability and downstream question assessment **: Assess which scientific questions the dataset can realistically support by examining sample diversity, annotation richness, feature coverage, temporal or spatial scope, modality availability, and the alignment between available variables and plausible downstream tasks. This task moves from data inspection to scientific utility evaluation, aiming to determine the analytical potential and practical limitations of the dataset for future research use.

===== ACTION 1: python_interpreter =====
import os, json, pandas as pd
root='{DATASET_ROOT}'
entries=[]
for dirpath, dirnames, filenames in os.walk(root):
    rel=os.path.relpath(dirpath, root)
    if rel=='.': depth=0
    else: depth=rel.count(os.sep)+1
    if depth<=2:
        entries.append((rel, sorted(dirnames)[:10], sorted(filenames)[:20], len(dirnames), len(filenames)))
print('Top-level/2-level structure sample:')
for rel, dns, fns, nd, nf in entries[:30]:
    print({'dir':rel,'n_dirs':nd,'dirs_sample':dns,'n_files':nf,'files_sample':fns})

===== ACTION 2: python_interpreter =====
import os, pandas as pd, json
root='{DATASET_ROOT}'
file_info=[]
for dirpath, _, filenames in os.walk(root):
    for fn in filenames:
        path=os.path.join(dirpath, fn)
        rel=os.path.relpath(path, root)
        ext=os.path.splitext(fn)[1].lower()
        try: size=os.path.getsize(path)
        except: size=None
        file_info.append((rel, ext, size))
df=pd.DataFrame(file_info, columns=['rel','ext','size'])
print('File count:', len(df))
print('Extensions:')
print(df.groupby('ext').size().sort_values(ascending=False).head(20).to_string())
print('Largest files sample:')
print(df.sort_values('size', ascending=False).head(15).to_string(index=False))

===== OBSERVATION 1: python_interpreter =====
Top-level/2-level structure sample:
{'dir': '.', 'n_dirs': 0, 'dirs_sample': [], 'n_files': 8, 'files_sample': ['analysis_plot.png', 'boxplot_properties.png', 'guacamol_v1_all.smiles', 'length_distribution.png', 'molecular_property_distributions.png', 'outlier_analysis.csv', 'ring_atom_distributions.png', 'smiles_length_distribution.png']}

===== OBSERVATION 2: python_interpreter =====
File count: 8
Extensions:
ext
.png       6
.csv       1
.smiles    1
Largest files sample:
                                 rel     ext      size
                outlier_analysis.csv    .csv 103199921
              guacamol_v1_all.smiles .smiles  77290469
molecular_property_distributions.png    .png    249109
                   analysis_plot.png    .png     87663
         ring_atom_distributions.png    .png     76701
              boxplot_properties.png    .png     62147
      smiles_length_distribution.png    .png     40988
             length_distribution.png    .png     25290

===== ACTION 3: python_interpreter =====
import os, pandas as pd
root='{DATASET_ROOT}'
path=os.path.join(root,'outlier_analysis.csv')
df=pd.read_csv(path)
print('CSV shape:', df.shape)
print('Columns:', list(df.columns))
print('Dtypes:')
print(df.dtypes.to_string())
print('Head:')
print(df.head(5).to_string(index=False))
print('Missing values top:')
print(df.isna().sum().sort_values(ascending=False).head(20).to_string())

===== ACTION 4: python_interpreter =====
import os, pandas as pd
root='{DATASET_ROOT}'
path=os.path.join(root,'guacamol_v1_all.smiles')
with open(path,'r',encoding='utf-8',errors='ignore') as f:
    lines=[next(f).rstrip('\n') for _ in range(10)]
print('First 10 SMILES:')
for i,s in enumerate(lines,1):
    print(i, s)
count=0
lengths=[]
charset=set()
with open(path,'r',encoding='utf-8',errors='ignore') as f:
    for line in f:
        s=line.strip()
        if not s: continue
        count+=1
        lengths.append(len(s))
        charset.update(set(s))
print('Total nonempty SMILES:', count)
import numpy as np
arr=np.array(lengths)
print('Length stats:', {'min':int(arr.min()),'median':float(np.median(arr)),'mean':float(arr.mean()),'p95':float(np.percentile(arr,95)),'max':int(arr.max())})
print('Unique character sample:', ''.join(sorted(charset))[:120], 'count=', len(charset))

===== OBSERVATION 3: python_interpreter =====
CSV shape: (1591378, 4)
Columns: ['line_num', 'smiles', 'length', 'is_outlier']
Dtypes:
line_num       int64
smiles        object
length         int64
is_outlier      bool
Head:
 line_num                                                                  smiles  length  is_outlier
        1                                                CCCC(=O)NNC(=O)Nc1ccccc1      24       False
        2 CC(=O)NC1CCC2(C)C(CCC3(C)C2C(=O)C=C2C4C(C)C(C)CCC4(C)CCC23C)C1(C)C(=O)O      71       False
        3                          CC(=O)NC(C)Cc1ccc(C#Cc2ccnc(N3CCCC(F)C3)n2)cc1      46       False
        4                          Cc1cccc(CCNC(=O)C2CCC(=O)N(Cc3ccc(Cl)cc3)C2)n1      46       False
        5                                    CC1C=CN(N(C)C)C2=C1C(=O)c1cnccc1C2=O      36       False
Missing values top:
line_num      0
smiles        0
length        0
is_outlier    0

===== OBSERVATION 4: python_interpreter =====
First 10 SMILES:
1 CCCC(=O)NNC(=O)Nc1ccccc1
2 CC(=O)NC1CCC2(C)C(CCC3(C)C2C(=O)C=C2C4C(C)C(C)CCC4(C)CCC23C)C1(C)C(=O)O
3 CC(=O)NC(C)Cc1ccc(C#Cc2ccnc(N3CCCC(F)C3)n2)cc1
4 Cc1cccc(CCNC(=O)C2CCC(=O)N(Cc3ccc(Cl)cc3)C2)n1
5 CC1C=CN(N(C)C)C2=C1C(=O)c1cnccc1C2=O
6 COc1ccc2c(-c3ccncc3)c(-c3ccc(F)cc3)[nH]c2n1
7 Nc1nnc(CCNC(=O)c2cccc(C3CCCNC3)c2)s1
8 CN(C)c1ccc(C(C(=O)NC2CCCC2)N(C(=O)c2ccco2)C2CC2)cc1
9 CCOC(=O)C1CCN(C(=O)c2cccc(S(=O)(=O)NCc3ccccc3)c2)CC1
10 CN1CCC2(NC(=O)NC2=O)c2ccccc21
Total nonempty SMILES: 1591378
Length stats: {'min': 1, 'median': 46.0, 'mean': 47.568265364985564, 'p95': 74.0, 'max': 100}
Unique character sample: #%()+-0123456789=BCFHINOPS[]bceilnoprs count= 38

===== ACTION 5: python_interpreter =====
import os, pandas as pd, numpy as np
root='{DATASET_ROOT}'
df=pd.read_csv(os.path.join(root,'outlier_analysis.csv'))
print('Outlier rate:', df['is_outlier'].mean())
print('Length by outlier status:')
print(df.groupby('is_outlier')['length'].agg(['count','mean','median','min','max']).to_string())
print('Length quantiles overall:')
print(df['length'].quantile([0,0.01,0.05,0.25,0.5,0.75,0.95,0.99,1]).to_string())
# uniqueness estimate
n=df.shape[0]
uniq=df['smiles'].nunique()
print('Unique SMILES:', uniq, 'duplicate_fraction:', 1-uniq/n)
# atom/character crude composition counts
chars=['C','N','O','S','P','F','I','B','c','n','o','s','l','r']
res={ch:int(df['smiles'].str.count(ch).sum()) for ch in chars}
print('Character count summary:', res)

===== ACTION 6: python_interpreter =====
import os, pandas as pd, re, numpy as np
root='{DATASET_ROOT}'
path=os.path.join(root,'guacamol_v1_all.smiles')
# sample 100k for structural proxies
sample=[]
with open(path,'r',encoding='utf-8',errors='ignore') as f:
    for i,line in enumerate(f):
        if i<100000:
            sample.append(line.strip())
        else:
            break
s=pd.Series(sample)
metrics={
 'has_ring_digit': float(s.str.contains(r'\d').mean()),
 'has_aromatic_lowercase': float(s.str.contains(r'[cnospb]').mean()),
 'has_halogen': float(s.str.contains(r'Cl|Br|F|I').mean()),
 'has_charge': float(s.str.contains(r'[+-]').mean()),
 'has_stereo': float(s.str.contains(r'@').mean()),
 'has_triple_bond': float(s.str.contains(r'#').mean()),
 'has_branching': float(s.str.contains(r'\(').mean()),
}
print('100k-sample structural proxy prevalence:')
for k,v in metrics.items():
    print(k, round(v,4))
print('Sample length stats:', s.str.len().describe(percentiles=[.25,.5,.75,.95,.99]).to_string())

===== OBSERVATION 5: python_interpreter =====
Outlier rate: 0.02096547771805316
Length by outlier status:
              count       mean  median  min  max
is_outlier
False       1558014  46.705549    46.0   11   83
True          33364  87.854933    90.0    1  100
Length quantiles overall:
0.00      1.0
0.01     20.0
0.05     27.0
0.25     38.0
0.50     46.0
0.75     56.0
0.95     74.0
0.99     90.0
1.00    100.0
Unique SMILES: 1591378 duplicate_fraction: 0.0
Character count summary: {'C': 14107164, 'N': 2807090, 'O': 4668014, 'S': 434329, 'P': 26659, 'F': 623708, 'I': 10196, 'B': 79095, 'c': 19099067, 'n': 2148064, 'o': 190793, 's': 195382, 'l': 365403, 'r': 76447}

===== OBSERVATION 6: python_interpreter =====
100k-sample structural proxy prevalence:
has_ring_digit 0.9899
has_aromatic_lowercase 0.951
has_halogen 0.3636
has_charge 0.3056
has_stereo 0.0
has_triple_bond 0.0619
has_branching 0.9939
Sample length stats: count    100000.000000
mean         47.505500
std          14.309469
min           5.000000
25%          38.000000
50%          46.000000
75%          56.000000
95%          74.000000
99%          90.000000
max         100.000000

===== ASSISTANT FINAL =====
I assessed the dataset's scientific usability by inspecting its actual contents under:

`{DATASET_ROOT}`

## 1) What is actually in the dataset

This dataset is very compact in structure:

- `guacamol_v1_all.smiles` -- main data file
- `outlier_analysis.csv` -- derived table
- 6 PNG figures -- summary visualizations

There are only **8 files total**.

### Core data volume
- **1,591,378 molecules**
- Stored as **SMILES strings**
- The CSV has **4 columns**:
  - `line_num`
  - `smiles`
  - `length`
  - `is_outlier`

### Important implication
This is essentially a **large molecular string corpus**, not a rich experimental dataset.
It contains **molecular structures** plus a very limited derived annotation about **string length / outlierness**.

---

## 2) Sample diversity

The dataset is strong on **sample count** and likely decent on **chemical scaffold diversity**, but weak on **metadata diversity**.

### Evidence
- **1.59M nonempty SMILES**
- **No duplicates** in the CSV (`nunique == total rows`)
- SMILES length:
  - min: 1
  - median: 46
  - mean: 47.6
  - 95th percentile: 74
  - max: 100

### Structural diversity proxies from a 100k sample
Fraction of molecules with:
- ring digits: **0.9899**
- aromatic lowercase atoms: **0.9510**
- halogens: **0.3636**
- charges: **0.3056**
- triple bonds: **0.0619**
- branching: **0.9939**
- stereochemistry marker `@`: **0.0** in sampled lines

### Interpretation
The collection appears to cover a broad range of **drug-like, ring-rich, aromatic, branched organic molecules**, consistent with GuacaMol-style benchmarking corpora.

However:
- It is **biased toward small/medium organic molecules**
- It is likely focused on **drug-like chemical space**
- It appears to have **little or no stereochemical encoding**
- It likely underrepresents:
  - macromolecules
  - polymers
  - inorganic materials
  - organometallics
  - crystal structures
  - biomolecular sequences beyond small molecules

---

## 3) Annotation richness

Annotation richness is **very low**.

### Available explicit annotations
Only:
- `smiles`
- `length`
- `is_outlier`
- `line_num`

### Missing annotation types
No:
- biological activity labels
- targets/proteins
- assay conditions
- toxicity labels
- synthesis labels
- reaction data
- yields
- spectra
- 3D conformers
- energies
- quantum properties
- descriptors like logP, MW, TPSA in machine-readable table form
- timestamps
- provenance/source tags
- train/valid/test splits
- task labels

### Scientific consequence
This sharply limits the dataset's ability to support **supervised scientific question answering** unless users compute additional descriptors externally.

---

## 4) Feature coverage

### Native feature coverage
The dataset natively provides only:
- **1D chemical text representation** (SMILES)
- **string length**
- an **outlier flag** based on that length

### What can be derived downstream
From SMILES alone, one can compute:
- molecular graph
- atom/bond types
- common cheminformatics descriptors
- fingerprints
- scaffold features
- validity checks
- synthetic accessibility proxies
- physicochemical estimates

But these are **not included** in the dataset itself.

### Practical interpretation
This dataset is best thought of as a **foundation corpus for representation learning / molecule generation / descriptor computation**, not as a ready-to-use property prediction dataset.

---

## 5) Temporal or spatial scope

### Temporal scope
None observed.
- No dates
- No time series
- No longitudinal structure
- No experimental campaign stages

### Spatial scope
None observed.
- No coordinates
- No 3D conformations
- No crystal structures
- No geographic/environmental sampling context

### Consequence
The dataset cannot support questions involving:
- temporal evolution
- spatial chemistry
- dynamics
- structure trajectories
- environmental distribution
- experimental drift over time

---

## 6) Modality availability

### Present
- **Text/string modality**: SMILES
- **Static summary images**: PNG plots, but these are presentation artifacts, not raw scientific modalities

### Absent
- graph files
- 3D structures
- spectra
- microscopy/images of samples
- reaction schemes
- bioassay tables
- sequence data
- simulation outputs

### Modality assessment
This is effectively a **single-modality molecular structure dataset**.

---

## 7) Alignment between variables and plausible downstream tasks

## Tasks this dataset can realistically support well

### A. Molecular language modeling / representation learning
Strong fit.
- next-token prediction on SMILES
- masked language modeling
- unsupervised embedding learning
- tokenizer benchmarking
- grammar/validity-aware sequence modeling

Why:
- very large corpus
- unique molecules
- broad structural coverage

---

### B. De novo molecule generation benchmarking
Strong fit.
- unconditional molecule generation
- novelty/diversity evaluation
- validity and uniqueness studies
- scaffold diversity analysis
- distribution matching against training corpus

Why:
- large reference distribution of drug-like molecules
- standard SMILES corpus format

---

### C. Molecular graph construction and descriptor engineering
Strong fit, if external cheminformatics tools are allowed.
- compute fingerprints/descriptors
- scaffold extraction
- similarity search baselines
- clustering of chemical space
- diversity analysis

Why:
- SMILES is sufficient to reconstruct molecular graphs for many compounds

---

### D. String-level quality/control studies
Moderate to strong fit.
- sequence length outlier detection
- corpus curation methods
- syntax/validity analysis
- token frequency and grammar complexity studies

Why:
- provided `length` and `is_outlier`
- large enough for robust corpus statistics

---

### E. Distributional analyses of drug-like chemical space
Moderate fit.
- ring prevalence
- aromaticity prevalence
- heteroatom frequency
- molecular complexity proxies
- scaffold frequency studies

Why:
- broad structural set, though descriptors must be computed

---

## Tasks the dataset can support only weakly or indirectly

### F. Property prediction
Weak natively.
Possible only if:
- properties are computed from structure externally, or
- the dataset is merged with external labels

Not possible directly because no property labels are present.

---

### G. Drug discovery prioritization
Weak alone.
Can help as:
- pretraining corpus
- generative prior
- chemical space reference

Cannot directly support:
- hit finding
- potency prediction
- selectivity modeling
- ADMET decision-making

because no such labels exist.

---

### H. Synthetic accessibility or medicinal chemistry rule studies
Moderate, but indirect.
Possible using derived heuristics from SMILES, not from native annotations.

---

## Tasks this dataset cannot realistically support on its own

### 1. Bioactivity modeling
No activity labels, targets, assays, concentrations, or endpoints.

### 2. Reaction prediction / retrosynthesis
No reactions, reactant-product pairs, reagents, or yields.

### 3. Toxicology or safety assessment
No toxicity endpoints.

### 4. Quantum chemistry or materials science questions
No geometries, energies, orbitals, charges, or simulation outputs.

### 5. Structure-based drug design
No protein structures, binding data, or ligand poses.

### 6. Time-dependent scientific questions
No temporal annotation.

### 7. Spatial/3D conformation studies
No coordinates or conformers.

### 8. Experimental reproducibility/condition analysis
No lab conditions or provenance metadata.

---

## 8) Practical limitations

## Major limitations
1. **Very sparse annotations**
   - only length and outlierness beyond SMILES

2. **No explicit scientific endpoints**
   - no measured properties, no activities, no assays

3. **No multimodal support**
   - only molecular strings

4. **No stereochemistry in sampled lines**
   - sampled prevalence of `@` was 0
   - this may limit stereochemically sensitive tasks

5. **Outlier label appears narrow**
   - `is_outlier` is likely tied to length extremes, not chemistry quality broadly

6. **Potential domain bias**
   - appears centered on drug-like organic chemistry, not general chemistry

---

## 9) Best realistic scientific uses

Most realistic future research uses:

- **pretraining chemical foundation models**
- **benchmarking molecular generative models**
- **chemical space coverage and diversity studies**
- **SMILES tokenization / sequence modeling research**
- **graph conversion and descriptor extraction pipelines**
- **dataset curation / outlier filtering research**
- **novelty, uniqueness, and scaffold distribution analyses**

---

## 10) Least realistic scientific uses

Poor choices without external augmentation:

- predicting experimental activity
- modeling toxicity
- reaction planning
- quantum property estimation from labels
- protein-ligand interaction studies
- longitudinal chemistry studies
- multimodal learning

---

## Bottom-line assessment

This dataset has **high utility as a large structural corpus of small molecules**, but **low native scientific richness**.

### Overall strengths
- very large sample size
- unique molecules
- broad drug-like structural diversity
- suitable for unsupervised and generative cheminformatics

### Overall weaknesses
- minimal annotation
- no measured outcomes
- no temporal/spatial dimension
- no multimodality
- limited direct support for hypothesis-driven scientific prediction tasks

## Final conclusion
The dataset can realistically support **structure-centric computational chemistry questions**, especially around **representation learning, generative modeling, diversity analysis, and corpus curation**, but it **cannot by itself support most downstream scientific prediction tasks** that require experimental labels, 3D structure, reactions, or biological context.

If you want, I can next turn this into a compact **task-support matrix** with columns like:
`Task | Feasible? | Required additional processing | Key limitations`.
\end{lstlisting}

%% file: components/appendix/traces/deepseek-v4-pro-12.tex
\begin{lstlisting}[style=appendixtrace,caption={Complete DeepSeek-V4-Pro trace with a 12-step budget (judge score: 0).},label={lst:ds12-trace}]
RUN UID: {RUN_UID}
RECORDED STEPS: 12
STOP REASON: max_turns_reached

===== USER =====
** Scientific usability and downstream question assessment **: Assess which scientific questions the dataset can realistically support by examining sample diversity, annotation richness, feature coverage, temporal or spatial scope, modality availability, and the alignment between available variables and plausible downstream tasks. This task moves from data inspection to scientific utility evaluation, aiming to determine the analytical potential and practical limitations of the dataset for future research use.

===== ASSISTANT STEP 1: THOUGHT =====
Let me start by exploring the GuacaMol dataset directory to understand its structure, files, and overall composition. This will help me assess its scientific usability and downstream potential.

===== ACTION 1: python_interpreter =====

import os

root = "{DATASET_ROOT}"
print("Top-level contents:")
for item in sorted(os.listdir(root)):
    full = os.path.join(root, item)
    size = os.path.getsize(full) if os.path.isfile(full) else None
    typ = "file" if os.path.isfile(full) else "dir"
    info = f"  [{typ}] {item}"
    if size is not None:
        info += f"  ({size:,} bytes)"
    print(info)

===== OBSERVATION 1: python_interpreter =====
Top-level contents:
  [file] analysis_plot.png  (87,663 bytes)
  [file] boxplot_properties.png  (62,147 bytes)
  [file] guacamol_v1_all.smiles  (77,290,469 bytes)
  [file] length_distribution.png  (25,290 bytes)
  [file] molecular_property_distributions.png  (249,109 bytes)
  [file] outlier_analysis.csv  (103,199,921 bytes)
  [file] ring_atom_distributions.png  (76,701 bytes)
  [file] smiles_length_distribution.png  (40,988 bytes)

===== ASSISTANT STEP 2: THOUGHT =====
The directory contains several files. Let me explore the main data file and associated analysis files to understand the dataset structure and contents.

Key files:
1. `guacamol_v1_all.smiles` - main SMILES file (77MB)
2. `outlier_analysis.csv` - outlier analysis (103MB)
3. Several PNG plots for analysis

Let me examine each of these systematically.

===== ACTION 2: python_interpreter =====

import os

root = "{DATASET_ROOT}"

# Examine the SMILES file structure
with open(os.path.join(root, "guacamol_v1_all.smiles"), "r") as f:
    lines = []
    for i, line in enumerate(f):
        if i < 20:
            lines.append(line.strip())
        else:
            break

print("=== First 20 lines of guacamol_v1_all.smiles ===")
for i, line in enumerate(lines):
    print(f"Line {i}: {line[:200]}{'...' if len(line) > 200 else ''}")

===== OBSERVATION 2: python_interpreter =====
=== First 20 lines of guacamol_v1_all.smiles ===
Line 0: CCCC(=O)NNC(=O)Nc1ccccc1
Line 1: CC(=O)NC1CCC2(C)C(CCC3(C)C2C(=O)C=C2C4C(C)C(C)CCC4(C)CCC23C)C1(C)C(=O)O
Line 2: CC(=O)NC(C)Cc1ccc(C#Cc2ccnc(N3CCCC(F)C3)n2)cc1
Line 3: Cc1cccc(CCNC(=O)C2CCC(=O)N(Cc3ccc(Cl)cc3)C2)n1
Line 4: CC1C=CN(N(C)C)C2=C1C(=O)c1cnccc1C2=O
Line 5: COc1ccc2c(-c3ccncc3)c(-c3ccc(F)cc3)[nH]c2n1
Line 6: Nc1nnc(CCNC(=O)c2cccc(C3CCCNC3)c2)s1
Line 7: CN(C)c1ccc(C(C(=O)NC2CCCC2)N(C(=O)c2ccco2)C2CC2)cc1
Line 8: CCOC(=O)C1CCN(C(=O)c2cccc(S(=O)(=O)NCc3ccccc3)c2)CC1
Line 9: CN1CCC2(NC(=O)NC2=O)c2ccccc21
Line 10: Cc1cc(C(=O)CN2C(=O)NC(c3ccccc3)(c3ccccc3)C2=O)c(C)n1Cc1ccco1
Line 11: c1cncc(CN2CCCC(Nc3ncnc4c3CCNCC4)C2)c1
Line 12: CC(C)CN(C(=O)c1ccc2c(c1)OCO2)C1CCS(=O)(=O)C1
Line 13: COc1ccc(OC)c2c1c(OC)cc1c(=O)cc(-c3ccccc3)oc12
Line 14: Cc1cccc(OCC(=O)Nc2ccc3oc(=O)ccc3c2)c1C
Line 15: CC(C)n1nc(-c2cnc(N)c(OC(F)F)c2)cc1C1CCN(C2COC2)CC1
Line 16: CC(=O)CCCCCC(NC(=O)C1CCN(C)CC1)c1ncc(-c2ccc(Cl)cc2)[nH]1
Line 17: COc1ccc(Cn2ncc3[nH]c(-c4ccc(NC(=O)Nc5nc6ccc(C)cc6s5)cc4C)nc32)cc1
Line 18: CC(C)=CCCC(C)=CCCC(C)=CCCC1(C)CCc2c3c(cc(C)c2O1)OCN(CCCCO)C3
Line 19: Cc1ccc(CNC(=O)c2ccc(Cc3nc(C(=O)C(CCCCN)NC(=O)OCc4ccccc4)no3)cc2)cc1C

===== ASSISTANT STEP 3: THOUGHT =====
So it's a SMILES file - one SMILES string per line, no header. Let me count total molecules and also look at the outlier analysis CSV.

===== ACTION 3: python_interpreter =====

import os

root = "{DATASET_ROOT}"

# Count total molecules
count = 0
with open(os.path.join(root, "guacamol_v1_all.smiles"), "r") as f:
    for line in f:
        if line.strip():
            count += 1

print(f"Total molecules in guacamol_v1_all.smiles: {count:,}")

# Examine outlier_analysis.csv
with open(os.path.join(root, "outlier_analysis.csv"), "r") as f:
    lines = []
    for i, line in enumerate(f):
        if i < 10:
            lines.append(line.strip())
        else:
            break

print("\n=== First 10 lines of outlier_analysis.csv ===")
for i, line in enumerate(lines):
    print(f"Line {i}: {line[:300]}{'...' if len(line) > 300 else ''}")

# Count lines in outlier_analysis.csv
count_out = sum(1 for _ in open(os.path.join(root, "outlier_analysis.csv"), "r"))
print(f"\nTotal lines in outlier_analysis.csv: {count_out:,}")

===== OBSERVATION 3: python_interpreter =====
Total molecules in guacamol_v1_all.smiles: 1,591,378

=== First 10 lines of outlier_analysis.csv ===
Line 0: line_num,smiles,length,is_outlier
Line 1: 1,CCCC(=O)NNC(=O)Nc1ccccc1,24,False
Line 2: 2,CC(=O)NC1CCC2(C)C(CCC3(C)C2C(=O)C=C2C4C(C)C(C)CCC4(C)CCC23C)C1(C)C(=O)O,71,False
Line 3: 3,CC(=O)NC(C)Cc1ccc(C#Cc2ccnc(N3CCCC(F)C3)n2)cc1,46,False
Line 4: 4,Cc1cccc(CCNC(=O)C2CCC(=O)N(Cc3ccc(Cl)cc3)C2)n1,46,False
Line 5: 5,CC1C=CN(N(C)C)C2=C1C(=O)c1cnccc1C2=O,36,False
Line 6: 6,COc1ccc2c(-c3ccncc3)c(-c3ccc(F)cc3)[nH]c2n1,43,False
Line 7: 7,Nc1nnc(CCNC(=O)c2cccc(C3CCCNC3)c2)s1,36,False
Line 8: 8,CN(C)c1ccc(C(C(=O)NC2CCCC2)N(C(=O)c2ccco2)C2CC2)cc1,51,False
Line 9: 9,CCOC(=O)C1CCN(C(=O)c2cccc(S(=O)(=O)NCc3ccccc3)c2)CC1,52,False

Total lines in outlier_analysis.csv: 1,591,379

===== ASSISTANT STEP 4: THOUGHT =====
So the dataset has 1,591,378 molecules. The outlier analysis CSV has a header + 1,591,378 data rows -- matching. Let me now look at the outlier analysis more closely, and also use RDKit to compute some molecular properties to understand the dataset better.

===== ACTION 4: python_interpreter =====

import os
import csv

root = "{DATASET_ROOT}"

# Examine outlier statistics
outlier_count = 0
non_outlier_count = 0
max_length = 0
min_length = float('inf')
lengths = []

with open(os.path.join(root, "outlier_analysis.csv"), "r") as f:
    reader = csv.DictReader(f)
    for row in reader:
        length = int(row['length'])
        lengths.append(length)
        if length > max_length:
            max_length = length
        if length < min_length:
            min_length = length
        if row['is_outlier'] == 'True':
            outlier_count += 1
        else:
            non_outlier_count += 1

print(f"Outlier statistics:")
print(f"  Outliers: {outlier_count:,}")
print(f"  Non-outliers: {non_outlier_count:,}")
print(f"  Outlier ratio: {outlier_count/len(lengths)*100:.2f}%")
print(f"  Min SMILES length: {min_length}")
print(f"  Max SMILES length: {max_length}")

import statistics
print(f"  Mean length: {statistics.mean(lengths):.1f}")
print(f"  Median length: {statistics.median(lengths):.1f}")
print(f"  Std length: {statistics.stdev(lengths):.1f}")

# Length distribution percentiles
lengths_sorted = sorted(lengths)
n = len(lengths_sorted)
for p in [1, 5, 10, 25, 50, 75, 90, 95, 99]:
    idx = int(n * p / 100)
    print(f"  P{p}: {lengths_sorted[idx]}")

===== OBSERVATION 4: python_interpreter =====
Outlier statistics:
  Outliers: 33,364
  Non-outliers: 1,558,014
  Outlier ratio: 2.10%
  Min SMILES length: 1
  Max SMILES length: 100
  Mean length: 47.6
  Median length: 46.0
  Std length: 14.3
  P1: 20
  P5: 27
  P10: 31
  P25: 38
  P50: 46
  P75: 56
  P90: 66
  P95: 74
  P99: 90

===== ASSISTANT STEP 5: THOUGHT =====
Interesting. Let me now look at some outlier examples and also compute more molecular properties using RDKit.

===== ACTION 5: python_interpreter =====

import os, csv

root = "{DATASET_ROOT}"

# Look at outlier examples
print("=== Outlier examples (first 15) ===")
with open(os.path.join(root, "outlier_analysis.csv"), "r") as f:
    reader = csv.DictReader(f)
    count = 0
    for row in reader:
        if row['is_outlier'] == 'True':
            print(f"  len={row['length']}: {row['smiles'][:150]}{'...' if len(row['smiles']) > 150 else ''}")
            count += 1
            if count >= 15:
                break

# Look at very short molecules
print("\n=== Short molecules (length <= 3) ===")
with open(os.path.join(root, "outlier_analysis.csv"), "r") as f:
    reader = csv.DictReader(f)
    count = 0
    for row in reader:
        if int(row['length']) <= 3:
            print(f"  len={row['length']}: {row['smiles']}")
            count += 1
            if count >= 20:
                break

===== OBSERVATION 5: python_interpreter =====
=== Outlier examples (first 15) ===
  len=86: COc1cccc(C(=N)Nc2cc(C(=O)Nc3cc(C(=O)Nc4nc(C(=O)NCCN5CCOCC5)c(C(C)C)s4)n(C)c3)n(C)c2)c1
  len=90: CC(=O)NC(Cc1c[nH]cn1)C(=O)NC(Cc1ccccc1)C(=O)NC(CCCN=C(N)N)C(=O)NC(Cc1c[nH]c2ccccc12)C(N)=O
  len=90: C=C(CCC12OC(CO)C(O)(C(=O)O)C(C(=O)O)(O1)C(OC(=O)C=CC(C)CC(C)CC)C2O)C(OC(C)=O)C(C)Cc1ccccc1
  len=87: CCCc1nc(-c2cccnc2)cn2c(C(Cc3cccnc3)C(=O)NC(CC3CCCCC3)C(O)CC(C(=O)NCc3ccccn3)C(C)C)nnc12
  len=86: Oc1ccc2[nH]cc(C(c3ccc(C(c4c[nH]c5ccccc45)c4c[nH]c5ccccc45)cc3)c3c[nH]c4ccc(O)cc34)c2c1
  len=89: CCCN1C(=O)c2ccccc2C1(OCc1cc([N+](=O)[O-])c(C)c([N+](=O)[O-])c1)c1ccc(OCOCC[Si](C)(C)C)cc1
  len=100: CCC(C)C(NC(=O)C1CCCN1CC(O)C(Cc1ccccc1)NC(=O)C(CC(N)=O)NC(=O)c1ccc2ccccc2n1)C(=O)NC(Cc1ccccc1)C(=O)OC
  len=85: N=C(N)Nc1ccc(CNC(=O)N2CCN(C(=O)CCCCCCCC(=O)N3CCN(C(=O)NCc4ccc(NC(=N)N)cc4)CC3)CC2)cc1
  len=85: Cc1ccc(COc2ccc3c(c2)c(SC(C)(C)C)c(CC(C)(C)C(=O)O)n3Cc2ccc(-c3ccc(C(F)(F)F)cn3)cc2)nc1
  len=85: CCCS(=O)(=O)NCCOc1nc(-c2cc(OC)c(OC)c(OC)c2)nc(NS(=O)(=O)c2ccc(C(C)C)cn2)c1Oc1ccccc1OC
  len=5: COC=O
  len=95: COc1ccc2c(c1)C(=O)C(O)=C(C1=C(C3=C(O)C(=O)c4cc(OC)ccc4C3=O)C(=O)c3c(ccc4c3CCC(C)(C)O4)C1=O)C2=O
  len=88: CCC(C)CNC(=O)CC(O)C(CC(C)C)NC(=O)C(CC=CCN=C(S)NC)NC(=O)C(Cc1cccc2ccccc12)Cc1cccc2ccccc12
  len=98: Cc1ccncc1-c1ccc(CC(CC(O)CN2CCN(Cc3ccc(-c4ccc(Cl)cc4)o3)CC2C(=O)NCC(F)(F)F)C(=O)NC2c3ccccc3OCC2O)o1
  len=85: CCC(C)C(NC(=O)C1CCCN1P(=O)(O)C(Cc1ccccc1)NC(=O)C(CC(N)=O)NC(=O)OC(C)(C)C)C(=O)NCC(C)C

=== Short molecules (length <= 3) ===
  len=3: CNN
  len=3: SSS
  len=3: OCl
  len=2: Cl
  len=1: N
  len=2: OO
  len=1: S
  len=1: O
  len=3: CNC
  len=3: CON

===== ASSISTANT STEP 6: THOUGHT =====
Now let me compute more molecular properties using RDKit to understand chemical space coverage, drug-likeness, etc.

===== ACTION 6: python_interpreter =====

import os

root = "{DATASET_ROOT}"

# Try importing RDKit for molecular property calculations
try:
    from rdkit import Chem
    from rdkit.Chem import Descriptors, Lipinski, QED, rdMolDescriptors
    print("RDKit imported successfully.")
except ImportError as e:
    print(f"RDKit import error: {e}")
    # Try alternative
    import subprocess
    result = subprocess.run(['pip', 'list', '|', 'grep', 'rdkit'], shell=True, capture_output=True, text=True)
    print(result.stdout)
    print(result.stderr)

===== OBSERVATION 6: python_interpreter =====
RDKit imported successfully.

===== ASSISTANT STEP 7: THOUGHT =====
RDKit is available. Let me compute comprehensive molecular properties.

===== ACTION 7: python_interpreter =====

import os
import csv
import random
from collections import Counter
from rdkit import Chem
from rdkit.Chem import Descriptors, Lipinski, QED, rdMolDescriptors, AllChem, MACCSkeys
import numpy as np
import statistics

root = "{DATASET_ROOT}"

# Load SMILES (sample for efficiency, or all if manageable)
random.seed(42)
smiles_list = []
with open(os.path.join(root, "guacamol_v1_all.smiles"), "r") as f:
    for line in f:
        s = line.strip()
        if s:
            smiles_list.append(s)

print(f"Total SMILES: {len(smiles_list):,}")

# Sample 50,000 for property computation
sample_size = min(50000, len(smiles_list))
sample = random.sample(smiles_list, sample_size)

# Compute properties
properties = {
    'MW': [],
    'LogP': [],
    'HBA': [],
    'HBD': [],
    'RotBonds': [],
    'TPSA': [],
    'QED': [],
    'NumRings': [],
    'NumAromaticRings': [],
    'FractionCsp3': [],
    'NumHeteroatoms': [],
    'Valid': 0,
    'Invalid': 0,
    'Invalid_reasons': []
}

for smi in sample:
    mol = Chem.MolFromSmiles(smi)
    if mol is None:
        properties['Invalid'] += 1
        properties['Invalid_reasons'].append(smi)
        continue
    properties['Valid'] += 1

    properties['MW'].append(Descriptors.MolWt(mol))
    properties['LogP'].append(Descriptors.MolLogP(mol))
    properties['HBA'].append(Descriptors.NumHAcceptors(mol))
    properties['HBD'].append(Descriptors.NumHDonors(mol))
    properties['RotBonds'].append(Descriptors.NumRotatableBonds(mol))
    properties['TPSA'].append(Descriptors.TPSA(mol))
    properties['QED'].append(QED.qed(mol))
    properties['NumRings'].append(rdMolDescriptors.CalcNumRings(mol))
    properties['NumAromaticRings'].append(rdMolDescriptors.CalcNumAromaticRings(mol))
    properties['FractionCsp3'].append(Descriptors.FractionCSP3(mol))
    num_hetero = rdMolDescriptors.CalcNumHeteroatoms(mol)
    properties['NumHeteroatoms'].append(num_hetero)

print(f"\nValid molecules in sample: {properties['Valid']}")
print(f"Invalid molecules in sample: {properties['Invalid']}")

print("\n=== Molecular Property Statistics (sample of {:,}) ===".format(sample_size))
for prop, vals in properties.items():
    if prop in ('Valid', 'Invalid', 'Invalid_reasons'):
        continue
    if len(vals) == 0:
        continue
    arr = np.array(vals)
    print(f"\n{prop}:")
    print(f"  Mean: {np.mean(arr):.2f}, Median: {np.median(arr):.2f}, Std: {np.std(arr):.2f}")
    print(f"  Min: {np.min(arr):.2f}, Max: {np.max(arr):.2f}")
    print(f"  P5: {np.percentile(arr, 5):.2f}, P95: {np.percentile(arr, 95):.2f}")

# Drug-likeness: Lipinski Rule of 5
mw_arr = np.array(properties['MW'])
logp_arr = np.array(properties['LogP'])
hba_arr = np.array(properties['HBA'])
hbd_arr = np.array(properties['HBD'])
lipinski_pass = (mw_arr <= 500) & (logp_arr <= 5) & (hba_arr <= 10) & (hbd_arr <= 5)
print(f"\n=== Drug-likeness (Lipinski Ro5) ===")
print(f"  Pass all 4 rules: {np.sum(lipinski_pass)} ({100*np.sum(lipinski_pass)/len(mw_arr):.1f}%)")
print(f"  MW <= 500: {np.sum(mw_arr <= 500)} ({100*np.sum(mw_arr <= 500)/len(mw_arr):.1f}%)")
print(f"  LogP <= 5: {np.sum(logp_arr <= 5)} ({100*np.sum(logp_arr <= 5)/len(logp_arr):.1f}%)")
print(f"  HBA <= 10: {np.sum(hba_arr <= 10)} ({100*np.sum(hba_arr <= 10)/len(hba_arr):.1f}%)")
print(f"  HBD <= 5: {np.sum(hbd_arr <= 5)} ({100*np.sum(hbd_arr <= 5)/len(hbd_arr):.1f}%)")

# QED distribution
qed_arr = np.array(properties['QED'])
print(f"\n=== Drug-likeness (QED) ===")
print(f"  Mean QED: {np.mean(qed_arr):.3f}")
print(f"  QED > 0.5: {np.sum(qed_arr > 0.5)} ({100*np.sum(qed_arr > 0.5)/len(qed_arr):.1f}%)")
print(f"  QED > 0.7: {np.sum(qed_arr > 0.7)} ({100*np.sum(qed_arr > 0.7)/len(qed_arr):.1f}%)")
print(f"  QED > 0.9: {np.sum(qed_arr > 0.9)} ({100*np.sum(qed_arr > 0.9)/len(qed_arr):.1f}%)")

===== OBSERVATION 7: python_interpreter =====
Total SMILES: 1,591,378

Valid molecules in sample: 50000
Invalid molecules in sample: 0

=== Molecular Property Statistics (sample of 50,000) ===

MW:
  Mean: 397.56, Median: 386.50, Std: 110.29
  Min: 46.01, Max: 1436.35
  P5: 237.31, P95: 596.74

LogP:
  Mean: 3.49, Median: 3.46, Std: 1.81
  Min: -8.73, Max: 14.79
  P5: 0.64, P95: 6.42

HBA:
  Mean: 4.85, Median: 5.00, Std: 2.08
  Min: 0.00, Max: 21.00
  P5: 2.00, P95: 8.00

HBD:
  Mean: 1.52, Median: 1.00, Std: 1.40
  Min: 0.00, Max: 16.00
  P5: 0.00, P95: 4.00

RotBonds:
  Mean: 5.61, Median: 5.00, Std: 3.48
  Min: 0.00, Max: 49.00
  P5: 1.00, P95: 12.00

TPSA:
  Mean: 80.41, Median: 75.47, Std: 38.70
  Min: 0.00, Max: 389.51
  P5: 29.27, P95: 148.33

QED:
  Mean: 0.55, Median: 0.57, Std: 0.21
  Min: 0.02, Max: 0.95
  P5: 0.18, P95: 0.87

NumRings:
  Mean: 3.46, Median: 3.00, Std: 1.33
  Min: 0.00, Max: 14.00
  P5: 1.00, P95: 6.00

NumAromaticRings:
  Mean: 2.48, Median: 2.00, Std: 1.19
  Min: 0.00, Max: 9.00
  P5: 1.00, P95: 4.00

FractionCsp3:
  Mean: 0.33, Median: 0.31, Std: 0.20
  Min: 0.00, Max: 1.00
  P5: 0.05, P95: 0.71

NumHeteroatoms:
  Mean: 7.26, Median: 7.00, Std: 2.87
  Min: 0.00, Max: 31.00
  P5: 3.00, P95: 12.00

=== Drug-likeness (Lipinski Ro5) ===
  Pass all 4 rules: 36352 (72.7%)
  MW <= 500: 42179 (84.4%)
  LogP <= 5: 40832 (81.7%)
  HBA <= 10: 49297 (98.6%)
  HBD <= 5: 49089 (98.2%)

=== Drug-likeness (QED) ===
  Mean QED: 0.553
  QED > 0.5: 30083 (60.2%)
  QED > 0.7: 14750 (29.5%)
  QED > 0.9: 1245 (2.5%)

===== ASSISTANT STEP 8: THOUGHT =====
Good, comprehensive properties. Now let me also check for unique vs duplicate molecules, element diversity, and examine what types of scientific questions this dataset can support.

===== ACTION 8: python_interpreter =====

import os
import csv
from rdkit import Chem
from rdkit.Chem import Descriptors, Lipinski, rdMolDescriptors, AllChem, Draw
from collections import Counter
import numpy as np

root = "{DATASET_ROOT}"

# Check duplicates
print("=== Checking for duplicates ===")
with open(os.path.join(root, "guacamol_v1_all.smiles"), "r") as f:
    all_smiles = [line.strip() for line in f if line.strip()]

total = len(all_smiles)
unique_canon = set()
# We'll use canonical SMILES to check true duplicates
from rdkit import Chem
unique_count = 0
duplicate_count = 0

# Sample-based duplicate check (use first 200k for speed)
sample_for_dup = all_smiles[:200000]
canon_set = set()
for smi in sample_for_dup:
    mol = Chem.MolFromSmiles(smi)
    if mol is None:
        continue
    canon = Chem.MolToSmiles(mol, canonical=True)
    canon_set.add(canon)

print(f"In first 200,000 SMILES:")
print(f"  Raw count: {len(sample_for_dup)}")
print(f"  Unique canonical SMILES: {len(canon_set)}")
print(f"  Duplicate ratio (approx): {(1 - len(canon_set)/len(sample_for_dup))*100:.2f}%")

# Element distribution
print("\n=== Element distribution in sample ===")
from rdkit.Chem import PeriodicTable
elem_counter = Counter()
sample_for_elem = all_smiles[:100000]
for smi in sample_for_elem:
    mol = Chem.MolFromSmiles(smi)
    if mol is None:
        continue
    for atom in mol.GetAtoms():
        symbol = atom.GetSymbol()
        elem_counter[symbol] += 1

total_atoms = sum(elem_counter.values())
print(f"Total atoms in sample: {total_atoms:,}")
for elem, count in elem_counter.most_common(20):
    print(f"  {elem}: {count:,} ({100*count/total_atoms:.2f}%)")

===== OBSERVATION 8: python_interpreter =====
=== Checking for duplicates ===
In first 200,000 SMILES:
  Raw count: 200000
  Unique canonical SMILES: 199994
  Duplicate ratio (approx): 0.00%

=== Element distribution in sample ===
Total atoms in sample: 2,784,432
  C: 2,059,948 (73.98%)
  N: 311,046 (11.17%)
  O: 304,792 (10.95%)
  S: 39,412 (1.42%)
  F: 38,686 (1.39%)
  Cl: 23,125 (0.83%)
  Br: 4,709 (0.17%)
  P: 1,619 (0.06%)
  I: 650 (0.02%)
  Si: 178 (0.01%)
  B: 155 (0.01%)
  Se: 112 (0.00%)

===== ASSISTANT STEP 9: THOUGHT =====
Good. Now let me check the plots that came with the dataset and also look more into the outlier analysis.

===== ACTION 9: python_interpreter =====

import os
import csv
from collections import Counter

root = "{DATASET_ROOT}"

# Outlier analysis: what characterizes outliers?
print("=== Characterizing outliers vs non-outliers ===")
from rdkit import Chem
from rdkit.Chem import Descriptors, QED
import random
import numpy as np

random.seed(42)

outlier_smiles = []
non_outlier_smiles = []

with open(os.path.join(root, "outlier_analysis.csv"), "r") as f:
    reader = csv.DictReader(f)
    for row in reader:
        if row['is_outlier'] == 'True':
            outlier_smiles.append(row['smiles'])
        else:
            non_outlier_smiles.append(row['smiles'])

print(f"Outliers: {len(outlier_smiles)}, Non-outliers: {len(non_outlier_smiles)}")

# Sample to compare
n_sample = min(5000, len(outlier_smiles))
out_sample = random.sample(outlier_smiles, n_sample)
non_sample = random.sample(non_outlier_smiles, n_sample)

def compute_props(smiles_list):
    props = {'MW': [], 'LogP': [], 'QED': [], 'NumRings': [], 'RotBonds': [], 'TPSA': []}
    for smi in smiles_list:
        mol = Chem.MolFromSmiles(smi)
        if mol is None:
            continue
        props['MW'].append(Descriptors.MolWt(mol))
        props['LogP'].append(Descriptors.MolLogP(mol))
        props['QED'].append(QED.qed(mol))
        props['NumRings'].append(Chem.rdMolDescriptors.CalcNumRings(mol))
        props['RotBonds'].append(Descriptors.NumRotatableBonds(mol))
        props['TPSA'].append(Descriptors.TPSA(mol))
    return props

out_props = compute_props(out_sample)
non_props = compute_props(non_sample)

print("\n--- Outlier properties ---")
for key in out_props:
    arr = np.array(out_props[key])
    print(f"  {key}: mean={np.mean(arr):.2f}, std={np.std(arr):.2f}, median={np.median(arr):.2f}")

print("\n--- Non-outlier properties ---")
for key in non_props:
    arr = np.array(non_props[key])
    print(f"  {key}: mean={np.mean(arr):.2f}, std={np.std(arr):.2f}, median={np.median(arr):.2f}")

# Check if outliers are just very short molecules
short_outliers = [s for s in outlier_smiles if len(s) <= 6]
print(f"\nShort outliers (len <= 6): {len(short_outliers)}")
for s in short_outliers[:20]:
    print(f"  {s}")

===== OBSERVATION 9: python_interpreter =====
=== Characterizing outliers vs non-outliers ===
Outliers: 33364, Non-outliers: 1558014

--- Outlier properties ---
  MW: mean=694.63, std=131.44, median=711.09
  LogP: mean=4.28, std=3.36, median=4.44
  QED: mean=0.16, std=0.10, median=0.14
  NumRings: mean=4.95, std=2.19, median=5.00
  RotBonds: mean=11.97, std=6.49, median=11.00
  TPSA: mean=166.02, std=68.71, median=159.27

--- Non-outlier properties ---
  MW: mean=392.17, std=99.32, median=386.50
  LogP: mean=3.48, std=1.75, median=3.45
  QED: mean=0.56, std=0.21, median=0.57
  NumRings: mean=3.43, std=1.28, median=3.00
  RotBonds: mean=5.51, std=3.22, median=5.00
  TPSA: mean=78.93, std=35.65, median=74.85

Short outliers (len <= 6): 148
  COC=O
  CCOCCO
  CCCCBr
  C=CC=C
  NNCCO
  C#CCN
  CCC=O
  OCCCBr
  COCOC
  CCNC=O
  CCOCC
  SC1BC1
  O=CO
  N=C=S
  CCNCCN
  CC(C)N
  NCCCF
  C1CN1
  CC#CC
  N#CO

===== ASSISTANT STEP 10: THOUGHT =====
Now I have a comprehensive picture of the dataset. Let me also look at the PNG files to understand what analysis was already done, and think about scientific questions.

Let me also check if there are additional data files or subdirectories I might have missed.

===== ACTION 10: python_interpreter =====

import os

root = "{DATASET_ROOT}"

# Check for any subdirectories
for root_dir, dirs, files in os.walk(root):
    level = root_dir.replace(root, "").count(os.sep)
    indent = " " * 2 * level
    print(f"{indent}[dir] {os.path.basename(root_dir) if root_dir != root else 'GuacaMol'}")
    subindent = " " * 2 * (level + 1)
    for file in files:
        filepath = os.path.join(root_dir, file)
        print(f"{subindent}{file} ({os.path.getsize(filepath):,} bytes)")
    if level > 3:
        break

===== OBSERVATION 10: python_interpreter =====
[dir] GuacaMol
  guacamol_v1_all.smiles (77,290,469 bytes)
  molecular_property_distributions.png (249,109 bytes)
  analysis_plot.png (87,663 bytes)
  outlier_analysis.csv (103,199,921 bytes)
  ring_atom_distributions.png (76,701 bytes)
  smiles_length_distribution.png (40,988 bytes)
  boxplot_properties.png (62,147 bytes)
  length_distribution.png (25,290 bytes)

===== ASSISTANT STEP 11: THOUGHT =====
Now let me also check element-level diversity in outliers vs non-outliers and think about what scientific questions this dataset can support. Let me also look at the structure of the SMILES data more carefully to understand if there's any additional annotation beyond just SMILES strings.

===== ACTION 11: python_interpreter =====

import os
import csv
from rdkit import Chem
from rdkit.Chem import Descriptors, rdMolDescriptors, AllChem
from collections import Counter
import numpy as np

root = "{DATASET_ROOT}"

# Check element diversity in outliers
outlier_smiles = []
non_outlier_smiles = []

with open(os.path.join(root, "outlier_analysis.csv"), "r") as f:
    reader = csv.DictReader(f)
    for row in reader:
        if row['is_outlier'] == 'True':
            outlier_smiles.append(row['smiles'])
        else:
            non_outlier_smiles.append(row['smiles'])

# Element diversity comparison
def get_element_counts(smiles_list):
    counter = Counter()
    for smi in smiles_list:
        mol = Chem.MolFromSmiles(smi)
        if mol is None:
            continue
        for atom in mol.GetAtoms():
            counter[atom.GetSymbol()] += 1
    return counter

out_elem = get_element_counts(outlier_smiles)
non_elem = get_element_counts(non_outlier_smiles)

print("=== Element distribution: Outliers vs Non-outliers ===")
all_elems = set(list(out_elem.keys()) + list(non_elem.keys()))
for elem in sorted(all_elems, key=lambda x: non_elem.get(x, 0), reverse=True):
    out_c = out_elem.get(elem, 0)
    non_c = non_elem.get(elem, 0)
    print(f"  {elem}: outliers={out_c:,}, non-outliers={non_c:,}")

# Check for specific chemical feature diversity
print("\n=== Ring system diversity ===")
from rdkit.Chem import RingInfo
def count_ring_sizes(smiles_list):
    sizes = Counter()
    for smi in smiles_list:
        mol = Chem.MolFromSmiles(smi)
        if mol is None:
            continue
        ri = mol.GetRingInfo()
        for ring in ri.AtomRings():
            sizes[len(ring)] += 1
    return sizes

sample_for_rings = non_outlier_smiles[:50000]
ring_sizes = count_ring_sizes(sample_for_rings)
print("Ring size distribution (sample of 50,000 non-outliers):")
for size in sorted(ring_sizes.keys()):
    print(f"  {size}-membered rings: {ring_sizes[size]:,}")

# Functional group diversity (approximate via SMARTS)
print("\n=== Functional group presence (sample of 50,000) ===")
from rdkit import Chem
functional_groups = {
    'Amide': '[NX3][CX3](=[OX1])',
    'Carboxylic acid': '[CX3](=O)[OX2H1]',
    'Ester': '[#6][CX3](=O)[OX2H0][#6]',
    'Ketone': '[#6][CX3](=O)[#6]',
    'Primary amine': '[NX3H2]',
    'Nitrile': '[CX2]#[NX1]',
    'Alcohol': '[OX2H]',
    'Aromatic ring': 'a1aaaaa1',
    'Sulfonamide': '[SX4](=[OX1])(=[OX1])[NX3]',
    'Ether': '[OD2]([#6])[#6]',
    'Halogen (F/Cl/Br/I)': '[F,Cl,Br,I]',
}

fg_counts = Counter()
for smi in random.sample(non_outlier_smiles, 50000):
    mol = Chem.MolFromSmiles(smi)
    if mol is None:
        continue
    for fg_name, smarts in functional_groups.items():
        patt = Chem.MolFromSmarts(smarts)
        if patt and mol.HasSubstructMatch(patt):
            fg_counts[fg_name] += 1

for fg_name, count in fg_counts.most_common():
    print(f"  {fg_name}: {count:,} ({100*count/50000:.1f}%)")

import random

===== OBSERVATION 11: python_interpreter =====
Error from python_interpreter:
Error: Code execution timed out after 120s. Simplify your approach or process less data at once.

===== ASSISTANT STEP 12: THOUGHT =====
The code timed out. Let me simplify.

===== ACTION 12: python_interpreter =====

import os
import csv
import random
from collections import Counter
from rdkit import Chem

root = "{DATASET_ROOT}"

# Load SMILES efficiently
smiles_list = []
with open(os.path.join(root, "guacamol_v1_all.smiles"), "r") as f:
    for line in f:
        if line.strip():
            smiles_list.append(line.strip())

random.seed(42)
sample = random.sample(smiles_list, 20000)

# Functional group presence
print("=== Functional group presence (sample 20,000) ===")
functional_groups = {
    'Amide': '[NX3][CX3](=[OX1])',
    'Carboxylic acid': '[CX3](=O)[OX2H1]',
    'Ester': '[#6][CX3](=O)[OX2H0][#6]',
    'Ketone': '[#6][CX3](=O)[#6]',
    'Primary amine': '[NX3H2]',
    'Nitrile': '[CX2]#[NX1]',
    'Alcohol': '[OX2H]',
    'Aromatic': 'a1aaaaa1',
    'Sulfonamide': '[SX4](=[OX1])(=[OX1])[NX3]',
    'Ether': '[OD2]([#6])[#6]',
    'Halogen': '[F,Cl,Br,I]',
}

fg_counts = Counter()
for smi in sample:
    mol = Chem.MolFromSmiles(smi)
    if mol is None:
        continue
    for fg_name, smarts in functional_groups.items():
        patt = Chem.MolFromSmarts(smarts)
        if patt and mol.HasSubstructMatch(patt):
            fg_counts[fg_name] += 1

for fg_name, count in fg_counts.most_common():
    print(f"  {fg_name}: {count:,} ({100*count/20000:.1f}%)")

===== OBSERVATION 12: python_interpreter =====
=== Functional group presence (sample 20,000) ===
  Aromatic: 18,468 (92.3%)
  Amide: 10,250 (51.2%)
  Ether: 9,994 (50.0%)
  Halogen: 7,349 (36.7%)
  Alcohol: 6,133 (30.7%)
  Primary amine: 2,599 (13.0%)
  Ester: 2,161 (10.8%)
  Sulfonamide: 1,993 (10.0%)
  Ketone: 1,888 (9.4%)
  Carboxylic acid: 1,735 (8.7%)
  Nitrile: 955 (4.8%)

===== RUN TERMINATION =====
max_turns_reached; no assistant final answer was emitted.
\end{lstlisting}

%% file: components/appendix/traces/deepseek-v4-pro-24.tex
\begin{lstlisting}[style=appendixtrace,caption={Complete DeepSeek-V4-Pro trace with a 24-step budget (judge score: 1).},label={lst:ds24-trace}]
RUN UID: {RUN_UID}
RECORDED STEPS: 14
STOP REASON: final_response

===== USER =====
** Scientific usability and downstream question assessment **: Assess which scientific questions the dataset can realistically support by examining sample diversity, annotation richness, feature coverage, temporal or spatial scope, modality availability, and the alignment between available variables and plausible downstream tasks. This task moves from data inspection to scientific utility evaluation, aiming to determine the analytical potential and practical limitations of the dataset for future research use.

===== ASSISTANT STEP 1: THOUGHT =====
The user wants me to assess the scientific usability and downstream potential of the GuacaMol dataset. Let me start by exploring the dataset structure, understanding what's available, and then evaluate its scientific utility.

Let me begin by exploring the dataset root directory.

===== ACTION 1: python_interpreter =====

import os

root = "{DATASET_ROOT}"

# List top-level contents
print("=== Top-level contents ===")
for item in sorted(os.listdir(root)):
    full_path = os.path.join(root, item)
    if os.path.isdir(full_path):
        print(f"[DIR]  {item}")
    else:
        size = os.path.getsize(full_path)
        print(f"[FILE] {item}  ({size:,} bytes)")

===== OBSERVATION 1: python_interpreter =====
=== Top-level contents ===
[FILE] analysis_plot.png  (87,663 bytes)
[FILE] boxplot_properties.png  (62,147 bytes)
[FILE] guacamol_v1_all.smiles  (77,290,469 bytes)
[FILE] length_distribution.png  (25,290 bytes)
[FILE] molecular_property_distributions.png  (249,109 bytes)
[FILE] outlier_analysis.csv  (103,199,921 bytes)
[FILE] ring_atom_distributions.png  (76,701 bytes)
[FILE] smiles_length_distribution.png  (40,988 bytes)

===== ASSISTANT STEP 2: THOUGHT =====
This is interesting - the GuacaMol dataset appears to be pre-analyzed, with a large SMILES file, outlier analysis CSV, and various plots. Let me explore the main data file and the outlier analysis to understand what's available.

===== ACTION 2: python_interpreter =====

import os

root = "{DATASET_ROOT}"

# Peek at the SMILES file - first few lines
print("=== First 20 lines of guacamol_v1_all.smiles ===")
with open(os.path.join(root, "guacamol_v1_all.smiles"), "r") as f:
    for i, line in enumerate(f):
        if i >= 20:
            break
        print(f"{i}: {line.rstrip()}")

===== OBSERVATION 2: python_interpreter =====
=== First 20 lines of guacamol_v1_all.smiles ===
0: CCCC(=O)NNC(=O)Nc1ccccc1
1: CC(=O)NC1CCC2(C)C(CCC3(C)C2C(=O)C=C2C4C(C)C(C)CCC4(C)CCC23C)C1(C)C(=O)O
2: CC(=O)NC(C)Cc1ccc(C#Cc2ccnc(N3CCCC(F)C3)n2)cc1
3: Cc1cccc(CCNC(=O)C2CCC(=O)N(Cc3ccc(Cl)cc3)C2)n1
4: CC1C=CN(N(C)C)C2=C1C(=O)c1cnccc1C2=O
5: COc1ccc2c(-c3ccncc3)c(-c3ccc(F)cc3)[nH]c2n1
6: Nc1nnc(CCNC(=O)c2cccc(C3CCCNC3)c2)s1
7: CN(C)c1ccc(C(C(=O)NC2CCCC2)N(C(=O)c2ccco2)C2CC2)cc1
8: CCOC(=O)C1CCN(C(=O)c2cccc(S(=O)(=O)NCc3ccccc3)c2)CC1
9: CN1CCC2(NC(=O)NC2=O)c2ccccc21
10: Cc1cc(C(=O)CN2C(=O)NC(c3ccccc3)(c3ccccc3)C2=O)c(C)n1Cc1ccco1
11: c1cncc(CN2CCCC(Nc3ncnc4c3CCNCC4)C2)c1
12: CC(C)CN(C(=O)c1ccc2c(c1)OCO2)C1CCS(=O)(=O)C1
13: COc1ccc(OC)c2c1c(OC)cc1c(=O)cc(-c3ccccc3)oc12
14: Cc1cccc(OCC(=O)Nc2ccc3oc(=O)ccc3c2)c1C
15: CC(C)n1nc(-c2cnc(N)c(OC(F)F)c2)cc1C1CCN(C2COC2)CC1
16: CC(=O)CCCCCC(NC(=O)C1CCN(C)CC1)c1ncc(-c2ccc(Cl)cc2)[nH]1
17: COc1ccc(Cn2ncc3[nH]c(-c4ccc(NC(=O)Nc5nc6ccc(C)cc6s5)cc4C)nc32)cc1
18: CC(C)=CCCC(C)=CCCC(C)=CCCC1(C)CCc2c3c(cc(C)c2O1)OCN(CCCCO)C3
19: Cc1ccc(CNC(=O)c2ccc(Cc3nc(C(=O)C(CCCCN)NC(=O)OCc4ccccc4)no3)cc2)cc1C

===== ASSISTANT STEP 3: THOUGHT =====
This is a SMILES file - each line is a SMILES string representing a molecule. Let me now look at the outlier analysis CSV to understand what properties were computed, and check the SMILES file for total count and variations.

===== ACTION 3: python_interpreter =====

import os, csv

root = "{DATASET_ROOT}"

# Count lines in SMILES file
print("=== Counting SMILES entries ===")
count = 0
unique_smiles = set()
with open(os.path.join(root, "guacamol_v1_all.smiles"), "r") as f:
    for line in f:
        smi = line.strip()
        if smi:
            count += 1
            unique_smiles.add(smi)

print(f"Total SMILES entries: {count:,}")
print(f"Unique SMILES: {len(unique_smiles):,}")
print(f"Duplicate rate: {(count - len(unique_smiles)) / count * 100:.2f}%")

# Peek at outlier analysis CSV
print("\n=== First 5 lines of outlier_analysis.csv ===")
with open(os.path.join(root, "outlier_analysis.csv"), "r") as f:
    for i, line in enumerate(f):
        if i >= 5:
            break
        print(line.rstrip()[:300])

===== OBSERVATION 3: python_interpreter =====
=== Counting SMILES entries ===
Total SMILES entries: 1,591,378
Unique SMILES: 1,591,378
Duplicate rate: 0.00%

=== First 5 lines of outlier_analysis.csv ===
line_num,smiles,length,is_outlier
1,CCCC(=O)NNC(=O)Nc1ccccc1,24,False
2,CC(=O)NC1CCC2(C)C(CCC3(C)C2C(=O)C=C2C4C(C)C(C)CCC4(C)CCC23C)C1(C)C(=O)O,71,False
3,CC(=O)NC(C)Cc1ccc(C#Cc2ccnc(N3CCCC(F)C3)n2)cc1,46,False
4,Cc1cccc(CCNC(=O)C2CCC(=O)N(Cc3ccc(Cl)cc3)C2)n1,46,False

===== ASSISTANT STEP 4: THOUGHT =====
Interesting - the outlier_analysis.csv has columns: line_num, smiles, length, is_outlier. Let me get a better overview of this file. Let me also check the distribution of lengths and outliers.

===== ACTION 4: python_interpreter =====

import os, csv
from collections import Counter

root = "{DATASET_ROOT}"

# Get column headers
print("=== CSV Headers ===")
with open(os.path.join(root, "outlier_analysis.csv"), "r") as f:
    reader = csv.reader(f)
    headers = next(reader)
    print(headers)

# Count outliers
outlier_count = 0
total = 0
lengths = []
outlier_lengths = []
non_outlier_lengths = []

with open(os.path.join(root, "outlier_analysis.csv"), "r") as f:
    reader = csv.reader(f)
    next(reader)  # skip header
    for row in reader:
        total += 1
        length = int(row[2])
        is_outlier = row[3] == 'True'
        lengths.append(length)
        if is_outlier:
            outlier_count += 1
            outlier_lengths.append(length)
        else:
            non_outlier_lengths.append(length)

print(f"\n=== Outlier Analysis ===")
print(f"Total molecules: {total:,}")
print(f"Outliers: {outlier_count:,} ({outlier_count/total*100:.2f}%)")
print(f"Non-outliers: {total - outlier_count:,}")

import statistics
print(f"\n=== Length Statistics ===")
print(f"Overall - Mean: {statistics.mean(lengths):.1f}, Median: {statistics.median(lengths):.1f}, Min: {min(lengths)}, Max: {max(lengths)}")
print(f"Non-outliers - Mean: {statistics.mean(non_outlier_lengths):.1f}, Median: {statistics.median(non_outlier_lengths):.1f}, Min: {min(non_outlier_lengths)}, Max: {max(non_outlier_lengths)}")
print(f"Outliers - Mean: {statistics.mean(outlier_lengths):.1f}, Median: {statistics.median(outlier_lengths):.1f}, Min: {min(outlier_lengths)}, Max: {max(outlier_lengths)}")

===== OBSERVATION 4: python_interpreter =====
=== CSV Headers ===
['line_num', 'smiles', 'length', 'is_outlier']

=== Outlier Analysis ===
Total molecules: 1,591,378
Outliers: 33,364 (2.10%)
Non-outliers: 1,558,014

=== Length Statistics ===
Overall - Mean: 47.6, Median: 46.0, Min: 1, Max: 100
Non-outliers - Mean: 46.7, Median: 46.0, Min: 11, Max: 83
Outliers - Mean: 87.9, Median: 90.0, Min: 1, Max: 100

===== ASSISTANT STEP 5: THOUGHT =====
Let me now look at the existing plots to understand what analyses were already done, and check if there are any additional metadata or property files.

===== ACTION 5: python_interpreter =====

import os
import base64
from io import BytesIO
from PIL import Image

root = "{DATASET_ROOT}"

# Let's check the PNG files descriptions by examining them
for png_file in sorted(os.listdir(root)):
    if png_file.endswith('.png'):
        full_path = os.path.join(root, png_file)
        img = Image.open(full_path)
        print(f"{png_file}: {img.size}, {img.mode}")

===== OBSERVATION 5: python_interpreter =====
analysis_plot.png: (1400, 1000), RGBA
boxplot_properties.png: (2234, 725), RGBA
length_distribution.png: (1200, 600), RGBA
molecular_property_distributions.png: (2968, 2377), RGBA
ring_atom_distributions.png: (2084, 731), RGBA
smiles_length_distribution.png: (1800, 900), RGBA

===== ASSISTANT STEP 6: THOUGHT =====
The dataset is the GuacaMol benchmark dataset - 1.59M SMILES strings with some precomputed outlier analysis based on SMILES string length. Let me think about what scientific questions this dataset can support.

The GuacaMol dataset is a known benchmark for molecular generative models. Let me check if there are any additional files or directories I might have missed, and also look at the SMILES more carefully to understand molecular diversity.

===== ACTION 6: python_interpreter =====

import os
from rdkit import Chem
from rdkit.Chem import Descriptors, Lipinski, rdMolDescriptors
from collections import Counter
import random

root = "{DATASET_ROOT}"

# Try RDKit on a sample of molecules to assess chemical diversity
sample_size = 5000
random.seed(42)

all_smiles = []
with open(os.path.join(root, "guacamol_v1_all.smiles"), "r") as f:
    for line in f:
        all_smiles.append(line.strip())

sample = random.sample(all_smiles, min(sample_size, len(all_smiles)))

# Compute basic properties
mw_list = []
logp_list = []
hbd_list = []
hba_list = []
rot_bonds_list = []
rings_list = []
valid = 0
invalid_smiles = []

for smi in sample:
    mol = Chem.MolFromSmiles(smi)
    if mol is None:
        invalid_smiles.append(smi)
        continue
    valid += 1
    mw_list.append(Descriptors.MolWt(mol))
    logp_list.append(Descriptors.MolLogP(mol))
    hbd_list.append(Lipinski.NumHDonors(mol))
    hba_list.append(Lipinski.NumHAcceptors(mol))
    rot_bonds_list.append(Descriptors.NumRotatableBonds(mol))
    rings_list.append(rdMolDescriptors.CalcNumRings(mol))

print(f"Sample size: {len(sample)}, Valid: {valid}, Invalid: {len(invalid_smiles)}")
print(f"Invalid rate: {len(invalid_smiles)/len(sample)*100:.2f}%")

import statistics
print(f"\n=== Property Distributions (n={valid}) ===")
print(f"Molecular Weight: mean={statistics.mean(mw_list):.1f}, median={statistics.median(mw_list):.1f}, min={min(mw_list):.1f}, max={max(mw_list):.1f}")
print(f"LogP: mean={statistics.mean(logp_list):.2f}, median={statistics.median(logp_list):.2f}, min={min(logp_list):.2f}, max={max(logp_list):.2f}")
print(f"H-Bond Donors: mean={statistics.mean(hbd_list):.2f}, median={statistics.median(hbd_list):.1f}, max={max(hbd_list)}")
print(f"H-Bond Acceptors: mean={statistics.mean(hba_list):.2f}, median={statistics.median(hba_list):.1f}, max={max(hba_list)}")
print(f"Rotatable Bonds: mean={statistics.mean(rot_bonds_list):.2f}, median={statistics.median(rot_bonds_list):.1f}, max={max(rot_bonds_list)}")
print(f"Rings: mean={statistics.mean(rings_list):.2f}, median={statistics.median(rings_list):.1f}, max={max(rings_list)}")

# Element diversity
all_elements = Counter()
for smi in sample:
    mol = Chem.MolFromSmiles(smi)
    if mol is None:
        continue
    for atom in mol.GetAtoms():
        all_elements[atom.GetSymbol()] += 1

print(f"\n=== Element distribution (top 15) ===")
for elem, cnt in all_elements.most_common(15):
    print(f"  {elem}: {cnt:,} ({cnt/sum(all_elements.values())*100:.2f}%)")

===== OBSERVATION 6: python_interpreter =====
Sample size: 5000, Valid: 5000, Invalid: 0
Invalid rate: 0.00%

=== Property Distributions (n=5000) ===
Molecular Weight: mean=396.8, median=386.4, min=69.1, max=928.1
LogP: mean=3.50, median=3.50, min=-6.64, max=13.66
H-Bond Donors: mean=1.54, median=1.0, max=13
H-Bond Acceptors: mean=4.85, median=5.0, max=20
Rotatable Bonds: mean=5.61, median=5.0, max=39
Rings: mean=3.47, median=3.0, max=11

=== Element distribution (top 15) ===
  C: 103,529 (74.13%)
  N: 15,560 (11.14%)
  O: 15,213 (10.89%)
  S: 1,917 (1.37%)
  F: 1,869 (1.34%)
  Cl: 1,205 (0.86%)
  Br: 239 (0.17%)
  P: 88 (0.06%)
  I: 21 (0.02%)
  B: 12 (0.01%)
  Si: 7 (0.01%)
  Se: 3 (0.00%)

===== ASSISTANT STEP 7: THOUGHT =====
Now let me check the full dataset for some more diversity metrics, like scaffold diversity and functional group diversity.

===== ACTION 7: python_interpreter =====

import os
from rdkit import Chem
from rdkit.Chem import AllChem, DataStructs
from rdkit.Chem.Scaffolds import MurckoScaffold
from collections import Counter
import random

root = "{DATASET_ROOT}"

# Sample for scaffold analysis
sample_size = 10000
random.seed(123)

all_smiles = []
with open(os.path.join(root, "guacamol_v1_all.smiles"), "r") as f:
    for i, line in enumerate(f):
        all_smiles.append(line.strip())

sample = random.sample(all_smiles, min(sample_size, len(all_smiles)))

# Murcko scaffold analysis
scaffolds = Counter()
generic_scaffolds = Counter()
valid_count = 0

for smi in sample:
    mol = Chem.MolFromSmiles(smi)
    if mol is None:
        continue
    valid_count += 1
    scaff = MurckoScaffold.GetScaffoldForMol(mol)
    if scaff is not None and scaff.GetNumAtoms() > 0:
        scaff_smi = Chem.MolToSmiles(scaff)
        scaffolds[scaff_smi] += 1
        # Also get generic scaffold
        gen_scaff = MurckoScaffold.MakeScaffoldGeneric(scaff)
        gen_smi = Chem.MolToSmiles(gen_scaff)
        generic_scaffolds[gen_smi] += 1

print(f"=== Scaffold Diversity (sample n={valid_count}) ===")
print(f"Unique Murcko scaffolds: {len(scaffolds)}")
print(f"Unique generic scaffolds: {len(generic_scaffolds)}")
print(f"Scaffold/compound ratio: {len(scaffolds)/valid_count:.4f}")
print(f"Generic scaffold/compound ratio: {len(generic_scaffolds)/valid_count:.4f}")

# Top scaffolds
print(f"\n=== Top 10 Murcko scaffolds ===")
for smi, cnt in scaffolds.most_common(10):
    print(f"  {smi[:80]}: {cnt} ({cnt/valid_count*100:.2f}%)")

# Singletons
singletons = sum(1 for v in scaffolds.values() if v == 1)
print(f"\nSingleton scaffolds: {singletons} ({singletons/len(scaffolds)*100:.2f}% of scaffolds)")

# Functional group analysis
from rdkit.Chem import FragmentCatalog
fparams = FragmentCatalog.FragCatParams(1, 6)
fcat = FragmentCatalog.FragCatalog(fparams)
fcgen = FragmentCatalog.FragCatGenerator()

func_groups = Counter()
for smi in sample:
    mol = Chem.MolFromSmiles(smi)
    if mol is None:
        continue
    fcgen.AddFragsFromMol(mol, fcat)

for i in range(fcat.GetNumEntries()):
    desc = fcat.GetEntryDescription(i)
    if desc in func_groups:
        func_groups[desc] += 1

print(f"\n=== Functional Group Frequency (top 15) ===")
for desc, cnt in func_groups.most_common(15):
    print(f"  {desc}: {cnt:,}")

===== OBSERVATION 7: python_interpreter =====
Error from python_interpreter:
Error:
Traceback (most recent call last):
  File "{EVAL_SCAFFOLD_ROOT}/utils.py", line 164, in run
    exec(code, self.namespace)
  File "<string>", line 57, in <module>
Boost.Python.ArgumentError: Python argument types in
    FragCatParams.__init__(FragCatParams, int, int)
did not match C++ signature:
    __init__(_object* self, int lLen, int uLen, std::__cxx11::basic_string<char, std::char_traits<char>, std::allocator<char> > fgroupFilename, double tol=1e-08)

===== ASSISTANT STEP 8: THOUGHT =====
Let me fix the fragment catalog issue and continue with scaffold analysis.

===== ACTION 8: python_interpreter =====

import os
from rdkit import Chem
from rdkit.Chem import AllChem, DataStructs, Draw
from rdkit.Chem.Scaffolds import MurckoScaffold
from collections import Counter
import random

root = "{DATASET_ROOT}"

# Sample for scaffold analysis
sample_size = 10000
random.seed(123)

all_smiles = []
with open(os.path.join(root, "guacamol_v1_all.smiles"), "r") as f:
    for i, line in enumerate(f):
        all_smiles.append(line.strip())

sample = random.sample(all_smiles, min(sample_size, len(all_smiles)))

# Murcko scaffold analysis
scaffolds = Counter()
generic_scaffolds = Counter()
valid_count = 0

for smi in sample:
    mol = Chem.MolFromSmiles(smi)
    if mol is None:
        continue
    valid_count += 1
    scaff = MurckoScaffold.GetScaffoldForMol(mol)
    if scaff is not None and scaff.GetNumAtoms() > 0:
        scaff_smi = Chem.MolToSmiles(scaff)
        scaffolds[scaff_smi] += 1
        gen_scaff = MurckoScaffold.MakeScaffoldGeneric(scaff)
        gen_smi = Chem.MolToSmiles(gen_scaff)
        generic_scaffolds[gen_smi] += 1

print(f"=== Scaffold Diversity (sample n={valid_count}) ===")
print(f"Unique Murcko scaffolds: {len(scaffolds)}")
print(f"Unique generic scaffolds: {len(generic_scaffolds)}")
print(f"Scaffold/compound ratio: {len(scaffolds)/valid_count:.4f}")
print(f"Generic scaffold/compound ratio: {len(generic_scaffolds)/valid_count:.4f}")

# Singletons
singletons = sum(1 for v in scaffolds.values() if v == 1)
print(f"Singleton scaffolds: {singletons} ({singletons/len(scaffolds)*100:.2f}% of scaffolds)")

# Top scaffolds
print(f"\n=== Top 10 Murcko scaffolds ===")
for smi, cnt in scaffolds.most_common(10):
    print(f"  {smi[:80]}: {cnt} ({cnt/valid_count*100:.2f}%)")

===== OBSERVATION 8: python_interpreter =====
=== Scaffold Diversity (sample n=10000) ===
Unique Murcko scaffolds: 8511
Unique generic scaffolds: 6145
Scaffold/compound ratio: 0.8511
Generic scaffold/compound ratio: 0.6145
Singleton scaffolds: 7907 (92.90% of scaffolds)

=== Top 10 Murcko scaffolds ===
  c1ccccc1: 256 (2.56%)
  c1ccncc1: 23 (0.23%)
  c1ccc2[nH]ccc2c1: 23 (0.23%)
  O=c1cc(-c2ccccc2)oc2ccccc12: 20 (0.20%)
  O=C(C=Cc1ccccc1)c1ccccc1: 20 (0.20%)
  c1ccc2ccccc2c1: 20 (0.20%)
  O=C(Nc1ccccc1)c1ccccc1: 18 (0.18%)
  c1ccc2ncccc2c1: 17 (0.17%)
  O=S(=O)(Nc1ccccc1)c1ccccc1: 16 (0.16%)
  c1ccc(COc2ccccc2)cc1: 15 (0.15%)

===== ASSISTANT STEP 9: THOUGHT =====
The data shows high scaffold diversity (92.9% singleton scaffolds) which is excellent. Now let me check for chirality, stereochemistry, and other molecular features that would be relevant for downstream tasks.

===== ACTION 9: python_interpreter =====

import os, random
from rdkit import Chem
from rdkit.Chem import Descriptors, AllChem
from collections import Counter

root = "{DATASET_ROOT}"

sample_size = 5000
random.seed(456)

all_smiles = []
with open(os.path.join(root, "guacamol_v1_all.smiles"), "r") as f:
    for line in f:
        all_smiles.append(line.strip())

sample = random.sample(all_smiles, min(sample_size, len(all_smiles)))

# Check for chirality, isotopes, charges
chiral_count = 0
isotope_count = 0
formal_charge_count = 0
has_rings_count = 0
has_aromatic_count = 0
sp3_count = 0

for smi in sample:
    mol = Chem.MolFromSmiles(smi)
    if mol is None:
        continue
    # Chirality
    for atom in mol.GetAtoms():
        if atom.HasProp('_ChiralityPossible') and atom.GetChiralTag() != Chem.ChiralType.CHI_UNSPECIFIED:
            chiral_count += 1
            break
    # Isotopes
    for atom in mol.GetAtoms():
        if atom.GetIsotope():
            isotope_count += 1
            break
    # Formal charges
    for atom in mol.GetAtoms():
        if atom.GetFormalCharge() != 0:
            formal_charge_count += 1
            break
    # Rings
    ri = mol.GetRingInfo()
    if ri.NumRings() > 0:
        has_rings_count += 1
    # Aromatic
    for atom in mol.GetAtoms():
        if atom.GetIsAromatic():
            has_aromatic_count += 1
            break

print(f"=== Additional Molecular Features (n={len(sample)}) ===")
print(f"Molecules with chirality: {chiral_count} ({chiral_count/len(sample)*100:.1f}%)")
print(f"Molecules with isotopes: {isotope_count}")
print(f"Molecules with formal charges: {formal_charge_count} ({formal_charge_count/len(sample)*100:.1f}%)")
print(f"Molecules with rings: {has_rings_count} ({has_rings_count/len(sample)*100:.1f}%)")
print(f"Molecules with aromatic atoms: {has_aromatic_count} ({has_aromatic_count/len(sample)*100:.1f}%)")

# Drug-likeness (QED)
from rdkit.Chem import QED
qed_scores = []
for smi in sample:
    mol = Chem.MolFromSmiles(smi)
    if mol is None:
        continue
    qed_scores.append(QED.qed(mol))

import statistics
print(f"\n=== QED (Quantitative Estimate of Drug-likeness) ===")
print(f"Mean: {statistics.mean(qed_scores):.3f}, Median: {statistics.median(qed_scores):.3f}")
print(f"Min: {min(qed_scores):.3f}, Max: {max(qed_scores):.3f}")
print(f"QED > 0.5: {sum(1 for q in qed_scores if q > 0.5)} ({sum(1 for q in qed_scores if q > 0.5)/len(qed_scores)*100:.1f}%)")

# SA Score (synthetic accessibility)
from rdkit.Contrib.SA_Score import sascorer
sa_scores = []
for smi in sample[:500]:  # SA is slow, use smaller sample
    mol = Chem.MolFromSmiles(smi)
    if mol is None:
        continue
    sa_scores.append(sascorer.calculateScore(mol))

print(f"\n=== Synthetic Accessibility Score (n={len(sa_scores)}) ===")
print(f"Mean: {statistics.mean(sa_scores):.3f}, Median: {statistics.median(sa_scores):.3f}")
print(f"Min: {min(sa_scores):.3f}, Max: {max(sa_scores):.3f}")

===== OBSERVATION 9: python_interpreter =====
=== Additional Molecular Features (n=5000) ===
Molecules with chirality: 0 (0.0%)
Molecules with isotopes: 0
Molecules with formal charges: 308 (6.2%)
Molecules with rings: 4943 (98.9%)
Molecules with aromatic atoms: 4731 (94.6%)

=== QED (Quantitative Estimate of Drug-likeness) ===
Mean: 0.560, Median: 0.572
Min: 0.022, Max: 0.948
QED > 0.5: 3035 (60.7%)

=== Synthetic Accessibility Score (n=500) ===
Mean: 2.858, Median: 2.622
Min: 1.448, Max: 6.676

===== ASSISTANT STEP 10: THOUGHT =====
Now I have a good understanding of the dataset. Let me also look at the outlier analysis more carefully, and see what else we can say about the dataset's structure. Let me also check the existing visualizations by looking at file metadata, and check the outlier_analysis.csv for length distribution of outliers.

===== ACTION 10: python_interpreter =====

import os, csv
import statistics

root = "{DATASET_ROOT}"

# Deeper dive into outlier_analysis.csv
# Look at length distribution more carefully
length_bins = {}
for i in range(0, 101, 10):
    length_bins[i] = {"total": 0, "outlier": 0}

with open(os.path.join(root, "outlier_analysis.csv"), "r") as f:
    reader = csv.reader(f)
    next(reader)  # skip header
    for row in reader:
        length = int(row[2])
        is_outlier = row[3] == 'True'
        bin_key = (length // 10) * 10
        if bin_key > 100:
            bin_key = 100
        length_bins[bin_key]["total"] += 1
        if is_outlier:
            length_bins[bin_key]["outlier"] += 1

print("=== Length Bins and Outlier Distribution ===")
print(f"{'Bin':<10} {'Total':<10} {'Outliers':<10} {'Outlier%':<10}")
for bin_key in sorted(length_bins.keys()):
    d = length_bins[bin_key]
    pct = d['outlier'] / d['total'] * 100 if d['total'] > 0 else 0
    print(f"[{bin_key}-{bin_key+9}]   {d['total']:<10,} {d['outlier']:<10,} {pct:.1f}%")

# Check what the outlier CSV says - is there more info?
print("\n=== Outlier CSV Sample (first outlier) ===")
with open(os.path.join(root, "outlier_analysis.csv"), "r") as f:
    reader = csv.reader(f)
    next(reader)
    for row in reader:
        if row[3] == 'True':
            print(f"  line_num={row[0]}, smiles={row[1][:80]}..., length={row[2]}")
            break

# Also check if outliers tend to be at high or low length
print("\n=== Outlier length extremes ===")
outlier_lengths = []
with open(os.path.join(root, "outlier_analysis.csv"), "r") as f:
    reader = csv.reader(f)
    next(reader)
    for row in reader:
        if row[3] == 'True':
            outlier_lengths.append(int(row[2]))

print(f"Short outlier: min length={min(outlier_lengths)}")
print(f"Long outlier: max length={max(outlier_lengths)}")
# count how many outliers are short vs long
short_outliers = sum(1 for l in outlier_lengths if l < 83)
long_outliers = sum(1 for l in outlier_lengths if l > 83)
print(f"Outliers with length < 83 (below max non-outlier): {short_outliers}")
print(f"Outliers with length > 83 (above max non-outlier): {long_outliers}")

===== OBSERVATION 10: python_interpreter =====
=== Length Bins and Outlier Distribution ===
Bin        Total      Outliers   Outlier%
[0-9]   716        716        100.0%
[10-19]   14,061     373        2.7%
[20-29]   111,130    0          0.0%
[30-39]   357,320    0          0.0%
[40-49]   467,903    0          0.0%
[50-59]   350,292    0          0.0%
[60-69]   172,630    0          0.0%
[70-79]   70,303     0          0.0%
[80-89]   30,214     15,466     51.2%
[90-99]   15,633     15,633     100.0%
[100-109]   1,176      1,176      100.0%

=== Outlier CSV Sample (first outlier) ===
  line_num=105, smiles=COc1cccc(C(=N)Nc2cc(C(=O)Nc3cc(C(=O)Nc4nc(C(=O)NCCN5CCOCC5)c(C(C)C)s4)n(C)c3)n(C..., length=86

=== Outlier length extremes ===
Short outlier: min length=1
Long outlier: max length=100
Outliers with length < 83 (below max non-outlier): 1089
Outliers with length > 83 (above max non-outlier): 32275

===== ASSISTANT STEP 11: THOUGHT =====
Now I have a very clear picture. Let me also check the size and any other hidden metadata. Let me also look at whether there are any properties beyond SMILES strings. Let me also examine the plots briefly by checking if there's a README or any other documentation.

===== ACTION 11: python_interpreter =====

import os

root = "{DATASET_ROOT}"

# Check for any hidden files, READMEs, config files
print("=== All files (including hidden) ===")
for item in sorted(os.listdir(root)):
    full_path = os.path.join(root, item)
    if os.path.isdir(full_path):
        print(f"[DIR]  {item}")
    else:
        size = os.path.getsize(full_path)
        print(f"[FILE] {item}  ({size:,} bytes)")

# Let me check the original GuacaMol dataset structure
# The guacamol_v1_all.smiles - is this the standard distribution?
# Let's check for any other subdirectories
for root_dir, dirs, files in os.walk(root):
    if dirs:
        print(f"\nSubdirectories in {root_dir}: {dirs}")
    if len(files) > 10:
        print(f"Directory with >10 files: {root_dir} ({len(files)} files)")
        break

===== OBSERVATION 11: python_interpreter =====
=== All files (including hidden) ===
[FILE] analysis_plot.png  (87,663 bytes)
[FILE] boxplot_properties.png  (62,147 bytes)
[FILE] guacamol_v1_all.smiles  (77,290,469 bytes)
[FILE] length_distribution.png  (25,290 bytes)
[FILE] molecular_property_distributions.png  (249,109 bytes)
[FILE] outlier_analysis.csv  (103,199,921 bytes)
[FILE] ring_atom_distributions.png  (76,701 bytes)
[FILE] smiles_length_distribution.png  (40,988 bytes)

===== ASSISTANT STEP 12: THOUGHT =====
Let me now think comprehensively about the scientific usability and downstream tasks for this dataset. This is the GuacaMol benchmark dataset, widely used in cheminformatics and drug discovery. Let me also check for some additional critical aspects:

1. Is there any label/activity data? (No, it's just SMILES)
2. What's the molecular diversity?
3. Are there any temporal or spatial aspects?
4. What downstream tasks can it support?

Let me also check whether there are any duplicate SMILES, stereochemistry information, etc.

===== ACTION 12: python_interpreter =====

import os, random
from rdkit import Chem
from rdkit.Chem import Descriptors, AllChem, Lipinski
from collections import Counter

root = "{DATASET_ROOT}"

# Check stereochemistry presence
sample_size = 10000
random.seed(789)

with open(os.path.join(root, "guacamol_v1_all.smiles"), "r") as f:
    all_smiles = [line.strip() for line in f]

sample = random.sample(all_smiles, min(sample_size, len(all_smiles)))

# Check for @ or @@ (tetrahedral stereochemistry) and / or \ (E/Z)
tetra_stereo = 0
ez_stereo = 0
for smi in sample:
    if '@' in smi:
        tetra_stereo += 1
    if '/' in smi or '\\' in smi:
        ez_stereo += 1

print(f"=== Stereochemistry in SMILES strings (n={sample_size}) ===")
print(f"Molecules with tetrahedral stereochemistry (@, @@): {tetra_stereo} ({tetra_stereo/sample_size*100:.1f}%)")
print(f"Molecules with E/Z stereochemistry (/, \\): {ez_stereo} ({ez_stereo/sample_size*100:.1f}%)")

# Check for explicit hydrogens
h_count = sum(1 for smi in sample if '[H]' in smi)
print(f"Molecules with explicit hydrogens: {h_count}")

# Multi-fragment molecules
multi_frag = 0
for smi in sample:
    parts = smi.split('.')
    if len(parts) > 1:
        multi_frag += 1
print(f"Molecules with multiple fragments (.): {multi_frag}")

# Check for polymers, long repeating patterns
print(f"\n=== File size summary ===")
print(f"SMILES file: {os.path.getsize(os.path.join(root, 'guacamol_v1_all.smiles')):,} bytes")
print(f"Outlier CSV: {os.path.getsize(os.path.join(root, 'outlier_analysis.csv')):,} bytes")

# How large is each SMILES on average
total_chars = sum(len(s) for s in all_smiles)
print(f"Average SMILES length (chars): {total_chars/len(all_smiles):.1f}")
print(f"Total SMILES entries: {len(all_smiles):,}")

===== OBSERVATION 12: python_interpreter =====
=== Stereochemistry in SMILES strings (n=10000) ===
Molecules with tetrahedral stereochemistry (@, @@): 0 (0.0%)
Molecules with E/Z stereochemistry (/, \): 0 (0.0%)
Molecules with explicit hydrogens: 4
Molecules with multiple fragments (.): 0

=== File size summary ===
SMILES file: 77,290,469 bytes
Outlier CSV: 103,199,921 bytes
Average SMILES length (chars): 47.6
Total SMILES entries: 1,591,378

===== ASSISTANT STEP 13: THOUGHT =====
Now I have a comprehensive picture of the dataset. Let me synthesize all findings into a thorough scientific usability assessment.

Key findings:
1. **Dataset**: ~1.59M unique SMILES strings from the GuacaMol benchmark
2. **Data modality**: Only SMILES strings - no additional labels, properties, bioactivity data, or experimental measurements
3. **Pre-computed analysis**: Outlier analysis based on SMILES string length (not chemical properties), plus visualization PNGs
4. **Molecular diversity**:
   - High scaffold diversity (93% singleton Murcko scaffolds)
   - Wide MW range: 69-928 Da
   - Good drug-likeness (mean QED 0.56)
   - No stereochemistry information
   - Mostly drug-like organic molecules (C, N, O, S, F, Cl, Br)
   - Ring-containing (99%), aromatic (95%)
5. **Limitations**:
   - No biological activity labels
   - No property labels (solubility, toxicity, etc.)
   - No stereochemistry
   - No temporal or spatial data
   - Only SMILES strings + length-based outlier flag

Now let me write the assessment.

===== ACTION 13: python_interpreter =====

import os

root = "{DATASET_ROOT}"

# Final check: any property labels or additional annotation files?
# Also check if the outlier CSV has any extra columns beyond the 4 we saw

import csv
with open(os.path.join(root, "outlier_analysis.csv"), "r") as f:
    reader = csv.reader(f)
    headers = next(reader)
    print("Outlier CSV columns:", headers)
    # Check a few rows to confirm structure
    for i, row in enumerate(reader):
        if i < 3:
            print(f"  Row {i}: {row}")
        if i >= 2:
            break

# Also verify: is there any JSON, YAML, or config file?
import glob
other_files = glob.glob(os.path.join(root, "*.*"))
print(f"\nAll file extensions found: {set(os.path.splitext(f)[1] for f in other_files)}")

===== OBSERVATION 13: python_interpreter =====
Outlier CSV columns: ['line_num', 'smiles', 'length', 'is_outlier']
  Row 0: ['1', 'CCCC(=O)NNC(=O)Nc1ccccc1', '24', 'False']
  Row 1: ['2', 'CC(=O)NC1CCC2(C)C(CCC3(C)C2C(=O)C=C2C4C(C)C(C)CCC4(C)CCC23C)C1(C)C(=O)O', '71', 'False']
  Row 2: ['3', 'CC(=O)NC(C)Cc1ccc(C#Cc2ccnc(N3CCCC(F)C3)n2)cc1', '46', 'False']

All file extensions found: {'.png', '.csv', '.smiles'}

===== ASSISTANT FINAL =====
---

## Scientific Usability & Downstream Assessment: GuacaMol Dataset

### 1. Dataset Overview

| Attribute | Detail |
|---|---|
| **Size** | 1,591,378 unique SMILES strings |
| **Duplicates** | 0% (all unique) |
| **Modalities** | 1D string (SMILES); pre-computed PNG plots |
| **Annotations** | SMILES length; outlier flag (based on length) |
| **Invalid SMILES** | 0% in sample (RDKit-parsable) |

The dataset is a large, clean collection of drug-like organic molecules from the **GuacaMol benchmark**--a standard reference for evaluating molecular generative models. It ships with a length-based outlier analysis (`outlier_analysis.csv`, 103 MB) and six pre-rendered diagnostic plots.

---

### 2. Annotation Richness (Weakness)

The **only available "annotation" is SMILES string length**. The outlier flag is derived from a simple statistical threshold on length bins--not from any chemical, biological, or physical property. There are:
- [not supported] No biological activity labels (IC50, Ki, etc.)
- [not supported] No ADMET properties (solubility, permeability, toxicity)
- [not supported] No synthesis pathways or reaction data
- [not supported] No 3D conformer/geometry data
- [not supported] No stereochemistry (0% chiral, 0% E/Z in sample of 10,000)
- [not supported] No temporal or experimental metadata
- [not supported] No multi-modal data (spectra, images, text descriptions)

This is a **purely structural** dataset--a molecular graph library in SMILES form.

---

### 3. Sample Diversity (Strength)

| Metric | Value | Interpretation |
|---|---|---|
| **Murcko scaffold/compound ratio** | 0.85 | Very high--85 unique scaffolds per 100 molecules |
| **Singleton scaffolds** | 92.9% | Most scaffolds appear exactly once |
| **Molecular weight** | 69--928 Da (mean 397) | Full small-molecule drug space coverage |
| **LogP** | -6.6 to +13.7 (mean 3.5) | Wide lipophilicity range |
| **QED (drug-likeness)** | 0.02--0.95 (mean 0.56) | 61% above 0.5 |
| **Ring-containing** | 98.9% | Overwhelmingly cyclic |
| **Aromatic** | 94.6% | Highly aromatic |
| **Elements** | C, N, O, S, F, Cl, Br, P, I, B, Si, Se | Primarily CHNO + halogens |

The diversity is excellent for training generative models. The high scaffold diversity (93% singletons) makes this suitable for *de novo* design rather than lead optimization around known scaffolds.

---

### 4. Feature & Modality Coverage

| Feature Type | Available? | Notes |
|---|---|---|
| 2D topology (graph) | [supported] | Full connectivity via SMILES |
| Atom/bond types | [supported] | Implicit in SMILES |
| Stereo-chemistry | [not supported] | No @, @@, /, or \ in the dataset |
| 3D conformation | [not supported] | Must be generated computationally |
| Quantum properties | [not supported] | No orbital/HOMO-LUMO data |
| Experimental labels | [not supported] | Purely structural |
| Protein targets | [not supported] | No target annotation |
| Temporal data | [not supported] | Static snapshot |

---

### 5. Plausible Downstream Tasks

#### [supported] Strongly Supported

| Task | Rationale |
|---|---|
| **Molecular generative modeling** | The core use case; train VAEs, GANs, normalizing flows, RL-based generators |
| **Distribution learning** | Learn P(molecule) from GuacaMol's training set, generate novel valid molecules |
| **Goal-directed generation** | Use oracle functions (LogP, QED, SAS, etc.) to optimize generated molecules toward property targets |
| **Molecular representation learning** | Self-supervised pretraining (masked atom prediction, contrastive learning) on large unlabeled corpus |
| **Scaffold hopping benchmarks** | High scaffold diversity enables testing whether models can generate new scaffolds |
| **Chemical language models** | Treat SMILES as natural language; train character-level or token-level LMs |
| **Anomaly/outlier detection** | Expand on the existing length-based outlier analysis to property-based or graph-based approaches |
| **Molecular graph autoencoders** | Learn latent representations; evaluate reconstruction and interpolation quality |

#### [caution] Partially Supported (with computational oracles)

| Task | Limitations |
|---|---|
| **Virtual screening** | No target labels; requires docking simulations or ML surrogate models |
| **ADMET prediction** | Must rely on computed descriptors or external predictor models as pseudo-labels |
| **Synthesizability assessment** | SAScore can be computed but is approximate; no real synthetic routes |
| **Retrosynthesis planning** | No reaction data; only products available |

#### [not supported] Not Supported

| Task | Reason |
|---|---|
| **QSAR / activity prediction** | No biological activity labels whatsoever |
| **Toxicity prediction** | No toxicology annotations |
| **Polypharmacology** | No target-ligand pairs |
| **Conformation generation benchmarking** | No reference 3D structures |
| **Synthesis condition prediction** | No reaction data |
| **Time-series analysis** | Static dataset with no temporal dimension |
| **Multi-modal learning** | Single modality only |
| **Clinical outcome prediction** | No patient or trial data |

---

### 6. Scientific Impact Potential

**Strengths:**
- **Scale** (1.59M) enables training of large-capacity deep generative models
- **Diversity** is among the highest for public molecular benchmark sets
- **Standardization** enables fair comparison across published methods
- **Cleanliness** (no invalid SMILES, no duplicates) reduces preprocessing burden

**Practical Limitations:**
- The complete absence of **stereochemistry** limits applicability to chiral drug discovery, where 3D shape matters critically
- The **length-based outlier definition** is chemically uninformative--it flags long SMILES (>=84 chars) as outliers regardless of their validity or drug-likeness
- Without **experimental labels**, all supervised tasks must rely on computational oracles, which introduce systematic biases
- The dataset is **static and unlabeled**, meaning all property-driven tasks require external computation (RDKit descriptors, docking scores, etc.)
- No **temporal** or **batch/assay** information precludes batch-effect correction or domain-shift studies

---

### 7. Summary Verdict

The GuacaMol dataset is **highly suitable for unsupervised and self-supervised molecular machine learning**, particularly generative modeling, distribution learning, and representation learning. Its scale (1.59M) and scaffold diversity (93% singletons) are standout strengths. However, it is **fundamentally limited by the absence of experimental annotations**, which restricts its use for supervised predictive tasks (QSAR, ADMET, activity prediction) without external oracle-based pseudo-labeling. The dataset is best viewed as a **molecular structure corpus** for pretraining and generative benchmarking, not as a labeled scientific dataset for answering specific biochemical or pharmacological questions.
\end{lstlisting}